\pdfoutput=1 
\documentclass[10pt,twocolumn,letterpaper]{article}

\usepackage{cvpr}              % To produce the CAMERA-READY version
\usepackage{graphicx}
\usepackage{amsmath}
\usepackage{amssymb}
\usepackage{booktabs}
\usepackage{threeparttable}
\usepackage{makecell}
\usepackage{multirow}
\usepackage[dvipsnames,table]{xcolor}
\usepackage{pifont}

\usepackage[accsupp]{axessibility} % Improves PDF readability for those with disabilities.
\usepackage{algorithm}
\usepackage{listings}

\usepackage[pagebackref,breaklinks,colorlinks]{hyperref}

\usepackage[capitalize]{cleveref}
\crefname{section}{Sec.}{Secs.}
\Crefname{section}{Section}{Sections}
\Crefname{table}{Table}{Tables}
\crefname{table}{Tab.}{Tabs.}

\newcommand{\cmark}{\ding{51}}%
\newcommand{\xmark}{\ding{55}}%
\definecolor{RowColor}{rgb}{0.95, 0.95, 1}

\def\cvprPaperID{2137} % *** Enter the CVPR Paper ID here
\def\confName{CVPR}
\def\confYear{2022}

\begin{document}

%%%%%%%%% TITLE - PLEASE UPDATE
\title{Surface Representation for Point Clouds}

\author{Haoxi Ran$^{1}$\thanks{corresponding author} \qquad Jun Liu$^2$ \qquad Chengjie Wang$^2$ \\
$^1$Northeastern University \qquad $^2$Tencent Youtu Lab \\ 
{\tt\normalsize ranhaoxi@gmail.com} \\
{\tt\normalsize \{juliusliu, jasoncjwang\}@tencent.com}
}
\maketitle

%%%%%%%%% ABSTRACT
\begin{abstract}

Most prior work represents the shapes of point clouds by coordinates. However, it is insufficient to describe the local geometry directly. In this paper, we present \textbf{RepSurf} (representative surfaces), a novel representation of point clouds to \textbf{explicitly} depict the very local structure. We explore two variants of RepSurf, Triangular RepSurf and Umbrella RepSurf inspired by triangle meshes and umbrella curvature in computer graphics. We compute the representations of RepSurf by predefined geometric priors after surface reconstruction. RepSurf can be a plug-and-play module for most point cloud models thanks to its free collaboration with irregular points. Based on a simple baseline of PointNet++ (SSG version), Umbrella RepSurf surpasses the previous state-of-the-art by a large margin for classification, segmentation and detection on various benchmarks in terms of performance and efficiency. With an increase of around \textbf{0.008M} number of parameters, \textbf{0.04G} FLOPs, and \textbf{1.12ms} inference time, our method achieves \textbf{94.7\%} (+0.5\%) on ModelNet40, and \textbf{84.6\%} (+1.8\%) on ScanObjectNN for classification, while \textbf{74.3\%} (+0.8\%) mIoU on S3DIS 6-fold, and \textbf{70.0\%} (+1.6\%) mIoU on ScanNet for segmentation. For detection, previous state-of-the-art detector with our RepSurf obtains \textbf{71.2\%} (+2.1\%) mAP$\mathit{_{25}}$, \textbf{54.8\%} (+2.0\%) mAP$\mathit{_{50}}$ on ScanNetV2, and \textbf{64.9\%} (+1.9\%) mAP$\mathit{_{25}}$, \textbf{47.7\%} (+2.5\%) mAP$\mathit{_{50}}$ on SUN RGB-D. Our lightweight Triangular RepSurf performs its excellence on these benchmarks as well. The code is publicly available at \url{https://github.com/hancyran/RepSurf}.

\end{abstract}

%%%%%%%%% BODY TEXT
\section{Introduction}

Learning from raw point clouds has drawn considerable attention for its advantages in various applications, like autonomous driving, augmented reality, and robotics. However, it can be difficult for the irregularity of point clouds. 

To handle irregular points, the pioneering work PointNet \cite{qi2017pointnet} adopts point-wise multi-layer perceptrons (MLP) to learn from points independently and utilizes a symmetric function to obtain the global information. PointNet++ \cite{qi2017pointnet++} further introduces \textit{set abstraction} (SA) to capture the local information of point clouds. However, both methods learn from standalone points and take no notice of local shape awareness \cite{liu2019relation}. 

Local shapes are vital for the learning of point clouds. To learn from the local structural information, some prior works learn from grids \cite{thomas2019kpconv, li2018pointcnn}, relations \cite{liu2019relation, ran2021learning}, or graphs \cite{wang2019dynamic, xu2020grid}. However, these methods learn from shapes indirectly by attaching more ingredients (like Euclidean distances, attention mechanism) or applying various transformations (like graph construction, voxelization). These may lead to complex preprocessing and significant computations. These sophisticated hand-crafted components learn from implicit local shape representations in general. We argue that it may lead to an omission of information when pre-defining the ingredients, or a loss of geometry during transformation.

%------------------------------------------------------------------------
\begin{figure}
\begin{center}
\scalebox{0.48}{\includegraphics[width=\textwidth]{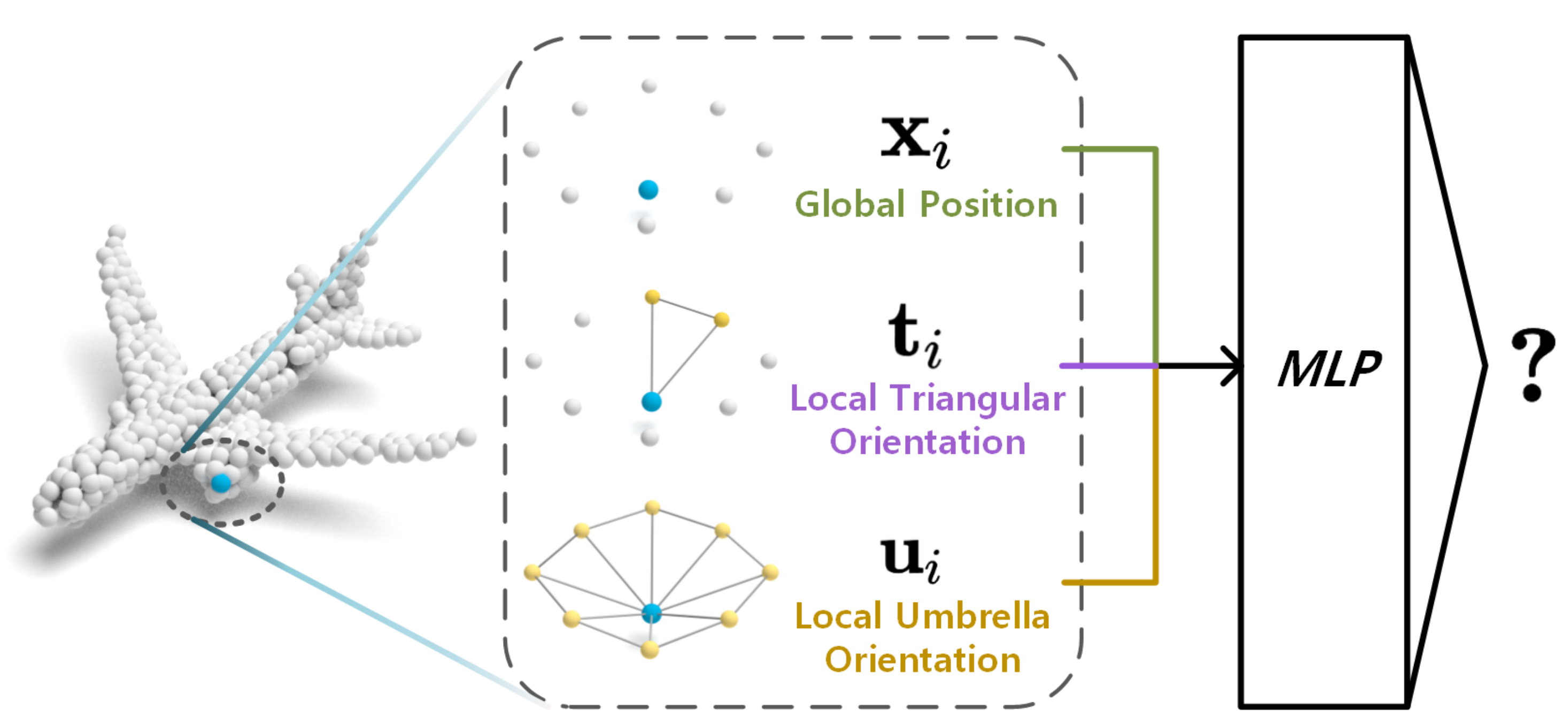}}
\end{center}
\vspace{-0.1in}
   \caption{An overview of point cloud classification with {\it RepSurf}. Given one point (blue) in the airplane point cloud, we indicate its global position by the coordinate $\mathbf{x}_i$. Different from the prior works, we further explicitly describe its local geometry through Triangular RepSurf $\mathbf{t}_i$ extracted from the reconstructed triangle or Umbrella RepSurf $\mathbf{u}_i$ learned from the reconstructed umbrella surface. By combining positional and geometric information, point representation can be more expressive. After concatenating $\mathbf{x}_i$ and $\mathbf{t}_i$/$\mathbf{u}_i$ as input, we predict the category of the point cloud via MLPs followed by a pooling operation.}
\label{fig:teaser}
\vspace{-0.2in}
\end{figure}
%------------------------------------------------------------------------

Taylor Series \cite{taylor1717methodus} expresses a local curve by derivatives. We simplify it by considering the second derivative only. Thus, we can roughly represent the local curve, or what we call the ``surface'' in 3D point clouds, by its corresponding tangent.

To this end, inspired by Taylor Series, we propose RepSurf (\textit{representative surfaces}) to explicitly represent the local shape of point clouds (shown in Fig.~\ref{fig:teaser}). To complement Cartesian coordinates in a point set with geometric information, we define RepSurf with three properties: discreteness, explicit locality, and curvature sensitivity. These properties allow RepSurf to express local geometry in free collaboration with irregular points. For a simple version of RepSurf, we propose Triangular RepSurf inspired by triangle meshes in computer graphics. We reconstruct a triangle for each point by querying its two neighbors and compute the triangle feature (i.e., normal vector, surface position, normalized coordinate) as RepSurf. To enlarge the perceptive field of RepSurf, we further propose Umbrella RepSurf inspired by umbrella curvature \cite{foorginejad2014umbrella}. Umbrella RepSurf can be an extension of Triangular RepSurf since it is computed from the triangles of an umbrella surface. Different from Triangular RepSurf, we reconstruct an umbrella surface after searching $K$ nearest neighbors and sorting the neighbors counterclockwise. For expressive representations, we feed the $K$ triangular features of an umbrella surface into a learnable transformation function followed by aggregation. Moreover, we present several delicate designs (i.e., polar auxiliary, channel de-differentiation) to further improve RepSurf.

Our key contributions are manifold:
\begin{itemize}
\item A novel triangle-based representation, Triangular RepSurf for point clouds.
\item A novel multi-surface representation, Umbrella RepSurf for point clouds.
\item A high-efficiency plug-and-play module based on RepSurf for point cloud models.
\item Our method achieves state-of-the-art on numerous point cloud benchmarks.
\end{itemize}

%%%%%%%%% Related Work
\section{Related Work}

%------------------------------------------------------------------------

\subsection{Learning on Point Clouds}
\textbf{Multi-view methods} 
\cite{feng2018gvcnn,guo2016multi,xie2016deepshape,han2018seqviews2seqlabels,qi2016volumetric,hamdi2021mvtn} or 
\textbf{voxel-based methods} 
\cite{wu20153d,maturana2015voxnet,gadelha2018multiresolution,choy20194d} 
describe 3D objects with multiple views (i.e., converting 3D shape to 2D images \cite{su2015multi} and latttice space \cite{su2018splatnet}) or by voxelization (Oc-tree based networks O-CNN \cite{wang2017cnn} and OctNet \cite{riegler2017octnet}, efficient submanifold sparse convolution \cite{graham20183d}). However, these transformation methods may lead to significant computations as well as a loss of shape information due to occlusion or lower resolution.

\textbf{Point-based methods} 
\cite{lang2020samplenet,fujiwara2020neural,le2020going,liu2020closer,nezhadarya2020adaptive,jiang2019hierarchical,hu2020randla,liu2019point2sequence} 
have recently attracted great attention to directly process point clouds. PointNet \cite{qi2017pointnet} learns from global information through multi-layer perceptrons and max-pooling operation. PointNet++ \cite{qi2017pointnet++} introduces set abstraction to capture the features from the local point sets, and farthest point sampling (FPS) to uniformly downsample between two set abstractions. Recent works explore local aggregator via 
convolutions \cite{xu2018spidercnn,wu2019pointconv,hermosilla2018monte,thomas2019kpconv,liu2019densepoint,zhang2019shellnet,lin2020convolution,lin2020fpconv,zhao2019pointweb,wang2018deep,wu2019pointconv,xu2021paconv,nie2021differentiable}, 
relations \cite{yan2020pointasnl,ran2021learning,zhao2021point,xiang2021walk}, and 
graphs \cite{wang2019dynamic,xu2020grid,zhou2021adaptive}. 
PointCNN \cite{li2018pointcnn} applies traditional convolution on point clouds after transforming neighboring points to the canonical order. RS-CNN \cite{liu2019relation} predefines geometric relations between points and their neighbors for local aggregation. DGCNN \cite{wang2019dynamic} computes the local graphs dynamically to extract geometric information. However, the methods are commonly based on some assumptions of implicit local geometry, which may result in missing geometric information in the input.

\subsection{Detection on Point Clouds}

Some early methods detect 3D objects by convolution after converting point clouds to 2D grids \cite{ku2018joint,liang2018deep,yang2018pixor,chen2017multi,xu2018multi} or 3D voxels \cite{song2016deep,zhou2018voxelnet}. Recent works focus on 3D detection of raw point clouds \cite{qi2018frustum,shi2019pointrcnn,shi2020pv,zhang2020h3dnet,chen2020hierarchical,qi2020imvotenet,misra2021end,cheng2021back,liu2021group}. VoteNet \cite{qi2019deep} adopts PointNet++ as the backbone for feature extraction and designs a component to group points corresponding to the voted centroids. \cite{liu2021group} removes the hand-crafted operation of grouping by introducing Transformers \cite{vaswani2017attention}.

%------------------------------------------------------------------------

\subsection{Graphics-related Surface Representation}
In computer graphics, triangle meshes are commonly adopted to represent 3D models. To obtain meshes from point clouds, previous works propose various methods for surface reconstruction. Ball-Pivoting Algorithm \cite{bernardini1999ball} forms a triangle if a specific-radius ball touches three points without containing other points. \cite{kazhdan2006poisson} defines the spatial Poisson formulation for surface reconstruction. 

Curvature can further present the local geometry on 3D point clouds. \cite{zhang2008curvature} estimates the local curvature of the point cloud surface by Least Square Fitting. \cite{foorginejad2014umbrella} constructs an umbrella surface based on the homogeneous neighbors and calculates the umbrella curvature through the neighbors' normal vectors and unit direction vectors.

%%%%%%%%% Methodology

\section{Surface Representation}

In this section, we first reveal the background for the design of our \textbf{Rep}resentative \textbf{Surf}aces (RepSurf) in Sec.~\ref{sec:background}. Secondly, we introduce several properties of RepSurf as inspiration in Sec.~\ref{sec:property}. Next, we propose two variants of RepSurf, Triangular and Umbrella RepSurf in Sec.~\ref{sec:triangular} and Sec.~\ref{sec:umbrella}, respectively. Finally, we implement RepSurf on PointNet++ (SSG version) and provide several exquisite designs to further improve the performance of RepSurf.

%------------------------------------------------------------------------

\subsection{Background}
\label{sec:background}

Local shapes are essential to represent point clouds. Prior works learn from shapes indirectly by utilizing extra ingredients or through different transformations. These operations may give some hints to express the local sets of point clouds, but cannot reflect the local shapes explicitly. We argue that the additional information leads to significant computations but contributes little to point cloud representations. Some may even cause the loss of geometric information. Therefore, we have to rethink on how to represent the local geometry.

We can describe a very local part centered on point $(t, f(t))$ of a 2D curve $f(\cdot)$ by Talyor series \cite{taylor1717methodus}:
\begin{equation}
f(x)=\sum_{n=0}^{\infty} \frac{f^{(n)}(t)}{n !}(x-t)^{n}, \: |x-t| < \epsilon
\end{equation}

\begin{figure}
\begin{center}
\scalebox{0.47}{\includegraphics[width=\textwidth]{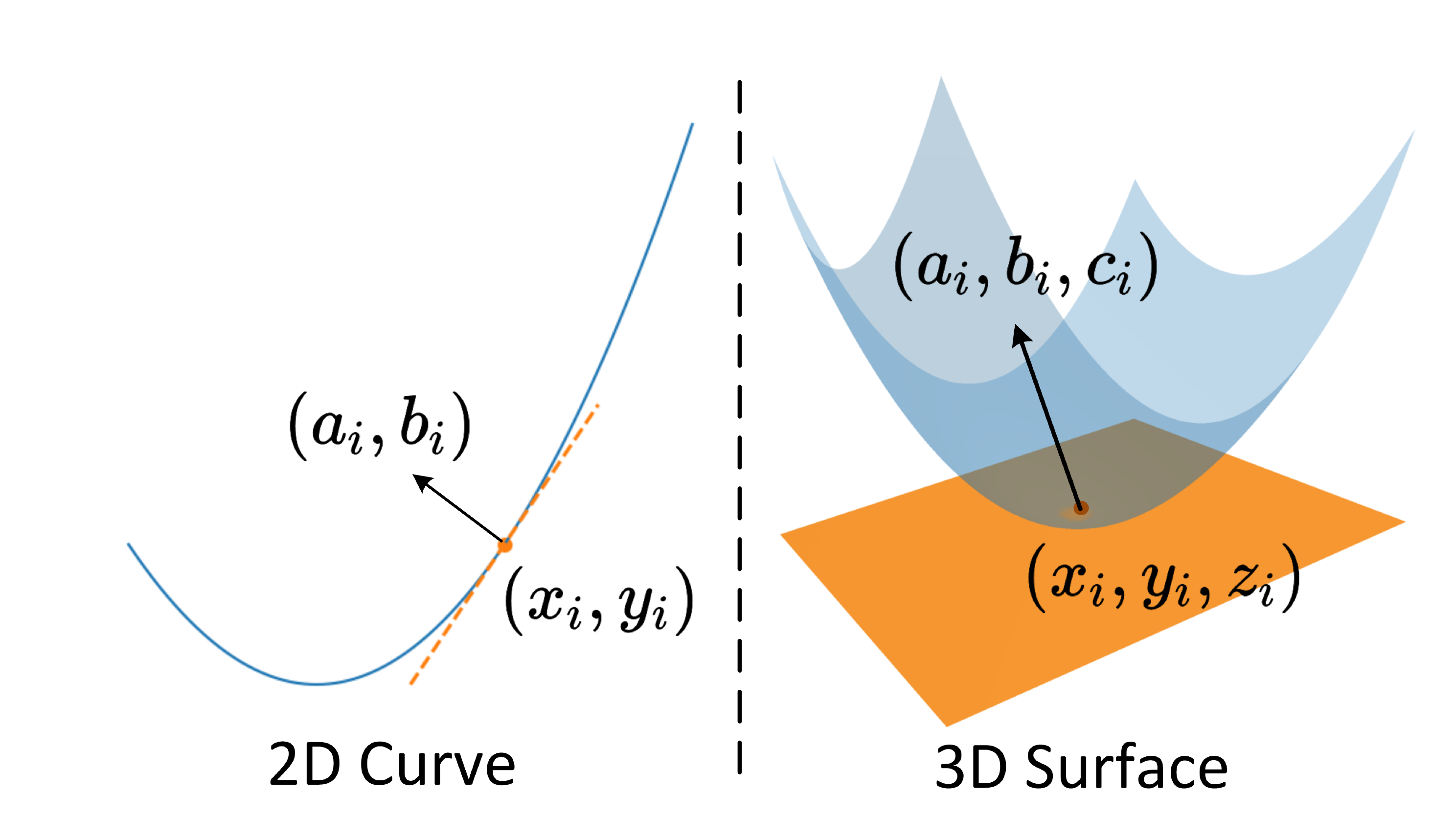}}
\end{center}
\vspace{-0.2in}
\caption{Local shape representation of a 2D curve (left) and a 3D surface (right) through the corresponding tangents.} 
\label{fig:tangent}
\end{figure}

To simplify the calculation, we approximate this equation by :
\begin{equation}
f(x) \simeq \underbrace{f(t)}_{\text {global position}}+\underbrace{f^{\prime}(t)}_{\text {local orientation}}(x-t),
\end{equation}
where $(t, f(t))$ is the global position on curve $f(\cdot)$, and the first derivative $f^{\prime}(t)$ can intuitively indicate the local orientation near point $(t, f(t))$. To further express the local curve (Fig.~\ref{fig:tangent} left), we represent the local orientation by its corresponding tangent:
\begin{align}
\begin{split}
a_i\left(x-x_{i}\right)+b_i\left(y-y_{i}\right)=0 \Rightarrow \\
a_ix+b_iy-(a_{i}x_{i}+b_{i}y_{i})=0,
\end{split}
\end{align} 
where $x_i=t$, $y_i=f(t)$, and $\frac{a_i}{b_i}=-f^{\prime}(a)$. $(a_i, b_i)$ is the normal vector of the tangent, where $a_i^2+b_i^2=1$. To conclude, a rough description of the local curve can be defined as:
\begin{equation}
\mathbf{c}_i = \left(x_i, y_i, a_i, b_i, a_{i}x_{i}+b_{i}y_{i}\right).
\end{equation}

%------------------------------------------------------------------------

\begin{figure}
\begin{center}
\scalebox{0.45}{\includegraphics[width=\textwidth]{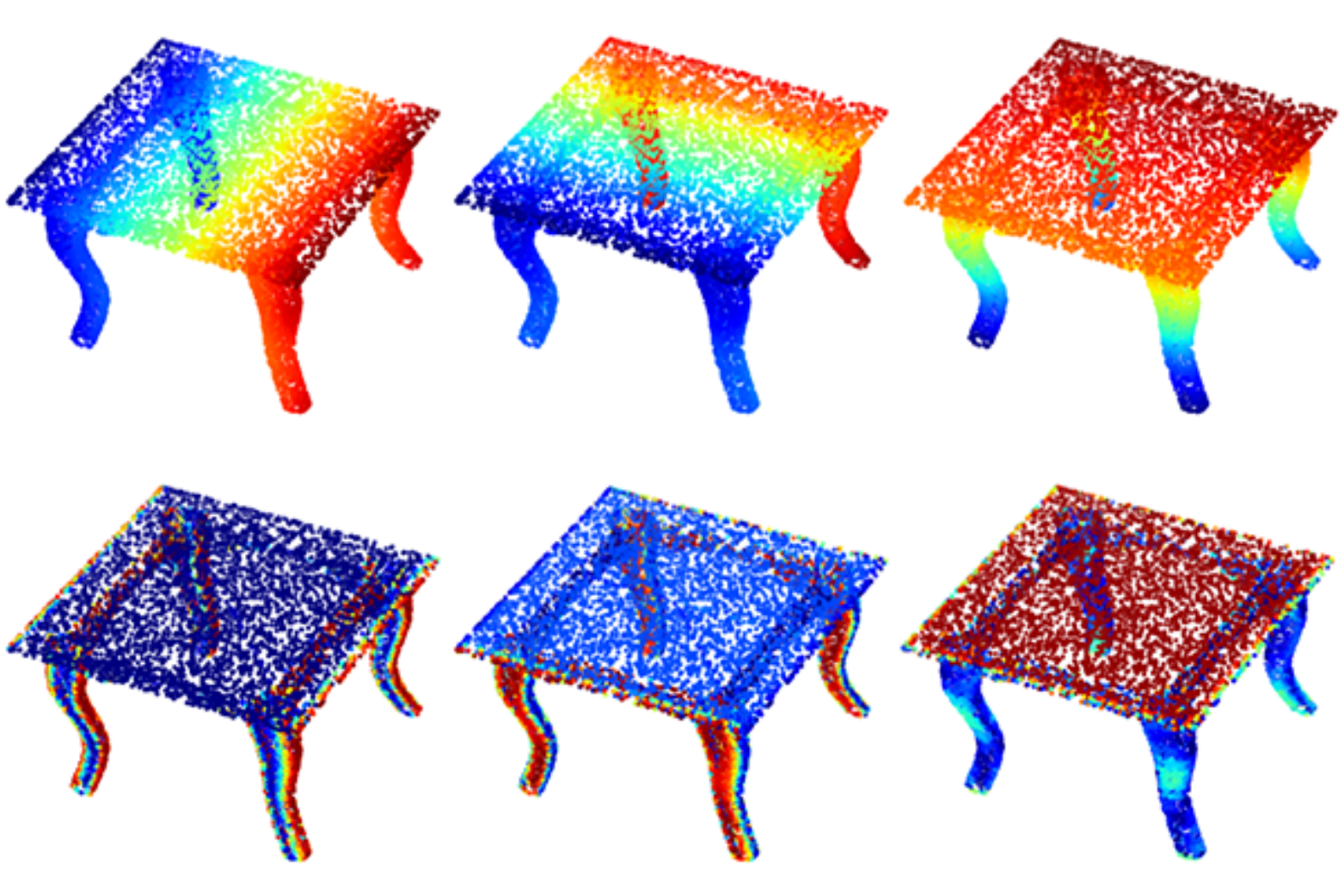}}
\end{center}
\vspace{-0.2in}
\caption{Visualization of a table on curvature sensitivity. We visualize a point cloud by the values of coordinates (above) and normals (below) in each of three dimensions. Intuitively, normal vectors can reflect the local shapes numerically to some extent.}
\label{fig:trivial}
\end{figure}

%------------------------------------------------------------------------

\subsection{Properties of RepSurf}
\label{sec:property}

PointNet \cite{qi2017pointnet} is inspired by three main properties of point sets in $\mathbb{R}^{N \times 3}$ from an Euclidean space: 1) unordered, 2) interaction among points and 3) invariance under transformations. It can handle the unordered point sets and alleviate the problem from rigid transformation. However, the ability to interact among points is still underexplored.

In 3D computer graphics, triangle meshes are a common representation of 3D models. Regularly, a triangle mesh consists of a set of triangles connected by their common edges or corners. Thus, triangles can flexibly present continuous and sophisticated 3D shapes for this characteristic. However, triangle meshes may not match the data structure of point clouds due to irregularity. A direct conversion from point cloud to triangle mesh may lead to significant computation as well as loss of point cloud characteristics (like flexibility from unorderness, scalability from the nature of sets). Therefore, we design our RepSurf inspired by the following properties:
\begin{itemize}
    \item \textbf{Discreteness}. Ideally, RepSurf should be a set to collaborate with the related point set. It means that each of $N$ points has a corresponding RepSurf feature.
    \item \textbf{Explicit Locality}. Unlike prior works describing local structure by learning (implicit locality), RepSurf shows the explicit locality of a part of point clouds numerically.
    \item \textbf{Curvature Sensitivity}. Coordinates can hardly depict the local shapes of 3D point clouds. RepSurf should be eligible to intuitively highlight edges and local shapes. An illustration is shown in Fig.~\ref{fig:trivial}.
\end{itemize}

%------------------------------------------------------------------------

\subsection{Triangular RepSurf}
\label{sec:triangular}

Denote a point set as $\mathbf{X}=\left\{\mathbf{x}_{1}, \ldots, \mathbf{x}_{n}\right\} \subseteq \mathbb{R}^{N \times 3}$. Analogous to a 2D curve in Sec~\ref{sec:background}, we define a 3D tangent surface (Fig.~\ref{fig:tangent} right) by point-normal equation. Given a normal vector $\mathbf{v}_i=(a_i, b_i, c_i)$ and a point $\mathbf{x}_{i}=(x_i, y_i, z_i)$, the surface can be defined as:
\begin{align}
\begin{split}
a_i\left(x-x_{i}\right)+b_i\left(y-y_{i}\right)+c_i\left(z-z_{i}\right)=0 \Rightarrow \\
a_ix+b_iy+c_iz-(a_{i}x_{i}+b_{i}y_{i}+c_{i}z_{i})=0.
\end{split}
\end{align} 

We define the surface position as $p_i=a_{i}x_{i}+b_{i}y_{i}+c_{i}z_{i}$, with the range of $[-\sqrt{3}r, \sqrt{3}r]$. $r$ means the edge length of a cube exactly covering the point set. For example, we utilize the normalized point clouds within the range of $[-1, 1]$ as input, so $r=1$ here. Note that $p_i$ can also express the directed distance between the origin and the surface. Then, we compute $\mathbf{v_i}$ by cross product. However, the computed $\mathbf{v_i}$ is unoriented --- $\mathbf{v_i}$ can be pointing either inside or outside of the surface. To handle this problem, prior works \cite{berger2017survey} adopt some time-costing methods. Considering efficiency, we simplify this case by keeping $a_i$ positive and augmenting the normals by instance-level random inverse with a probability of 50\%. Thus, we define Triangular RepSurf as:
\begin{equation}
\mathbf{t}_i = \left(a_i, b_i, c_i, p_i\right).
\end{equation}

We define a set of Triangular RepSurf as $\mathbf{T}=\left\{\mathbf{t}_{1}, \ldots, \mathbf{t}_{n}\right\} \subseteq \mathbb{R}^{N \times 4}$. To feed point clouds into models, we replace $\mathbf{X}$ with our re-computed centroids $\mathbf{X}'$ of the triangles. Then the input can be the concatenation of $\mathbf{X}'$ and $\mathbf{T}$. A simple illustration and the implementation of Triangular RepSurf is presented in Fig.~\ref{fig:teaser} and Algorithm \ref{alg:triangular}, respectively.

%##########################################Code of Triangular RepSurf#####################################
\begin{algorithm}[t]
\caption{\small Pytorch-Style Pseudocode of \textbf{Triangular RepSurf}}
\label{alg:triangular}
\definecolor{codeblue}{rgb}{0.25,0.5,0.7}
\vspace{-4pt}
\lstset{
  backgroundcolor=\color{white},
  basicstyle=\fontsize{7.6pt}{7.6pt}\ttfamily\selectfont,
  columns=fullflexible,
  breaklines=true,
  captionpos=b,
  commentstyle=\fontsize{10pt}{10pt}\color{codeblue},
  keywordstyle=\fontsize{10pt}{10pt},
}
\begin{lstlisting}[language=python]
# B: batch size, N: number of points
# points: coordinates of a point set
pairs = kNN(points, k=2)-points # [B,N,2,3]
centroids = mean(pairs, dim=2) # [B,N,3]
normals = cross_product(pairs) # [B,N,3]
normals = normals/norm(normals, dim=-1) # [B,N,3]
pos_mask = (normals[..., 0]>0)*2-1 # [B,N,1]
normals = normals*pos_mask # [B,N,3]
normals = random_inverse(normals) # [B,N,3]
positions = sum(normals*centroids, dim=2)/sqrt(3) 
       # [B,N,1]
out = concat([centroids, normals, positions], dim=2) 
       # [B,N,7]
return out
\end{lstlisting}
\vspace{-3pt}
\label{alg:pseudocode}
\end{algorithm}

%##################################################################################################

%#######################################Code of Umbrella RepSurf#########################################

\begin{algorithm}[t]
\caption{\small Pytorch-Style Pseudocode of \textbf{Umbrella RepSurf}}
\label{alg:umbrella}
\definecolor{codeblue}{rgb}{0.25,0.5,0.7}
\vspace{-4pt}
\lstset{
  backgroundcolor=\color{white},
  basicstyle=\fontsize{7.6pt}{7.6pt}\ttfamily\selectfont,
  columns=fullflexible,
  breaklines=true,
  captionpos=b,
  commentstyle=\fontsize{10pt}{10pt}\color{codeblue},
  keywordstyle=\fontsize{10pt}{10pt},
}
\begin{lstlisting}[language=python]
# B: batch size, N: number of points
# K: number of neighbors, C: output channels
# points: coordinates of a point set
neighbors = kNN(points, k=K)-points # [B,N,K,3]
edges = sort_by_clock(neighbors) # [B,N,K,3]
edges = unsqueeze(neighbors, dim=-2) # [B,N,K,1,3]
pairs = concat([edges, edges.roll(-1, 2)], dim=-2)
       # [B,N,K,2,3]
centroids = mean(pairs, dim=3) # [B,N,K,3]
normals = cross_product(pairs) # [B,N,K,3]
normals = normals/norm(normals, dim=-1) # [B,N,K,3]
pos_mask = (normals[..., 0, 0]>0)*2-1 # [B,N,1,1]
normals = normals*pos_mask # [B,N,K,3]
normals = random_inverse(normals) # [B,N,K,3]
positions = sum(normals*centroids, dim=3)/sqrt(3) 
       # [B,N,K,1]
features = concat([centroids, normals, positions], dim=2) # [B,N,K,7]
features = MLPs(features, out_channel=C) #[B,N,K,C]
features = pooling(features, dim=2) # [B,N,C]
out = concat([centroids, features], dim=2) 
       # [B,N,3+C]
return out
\end{lstlisting}
\vspace{-3pt}
\label{alg:pseudocode}
\end{algorithm}

%##################################################################################################

\subsection{Umbrella RepSurf}
\label{sec:umbrella}

Triangular RepSurf is a lightweight method to represent the local geometry of a point cloud. However, due to limited perceptive field, it may also lead to unstable local representations. To handle this drawback, we expand the perceptive field by proposing Umbrella RepSurf inspired by umbrella curvature \cite{foorginejad2014umbrella}. 

\begin{figure}
\begin{center}
\scalebox{0.45}{\includegraphics[width=\textwidth]{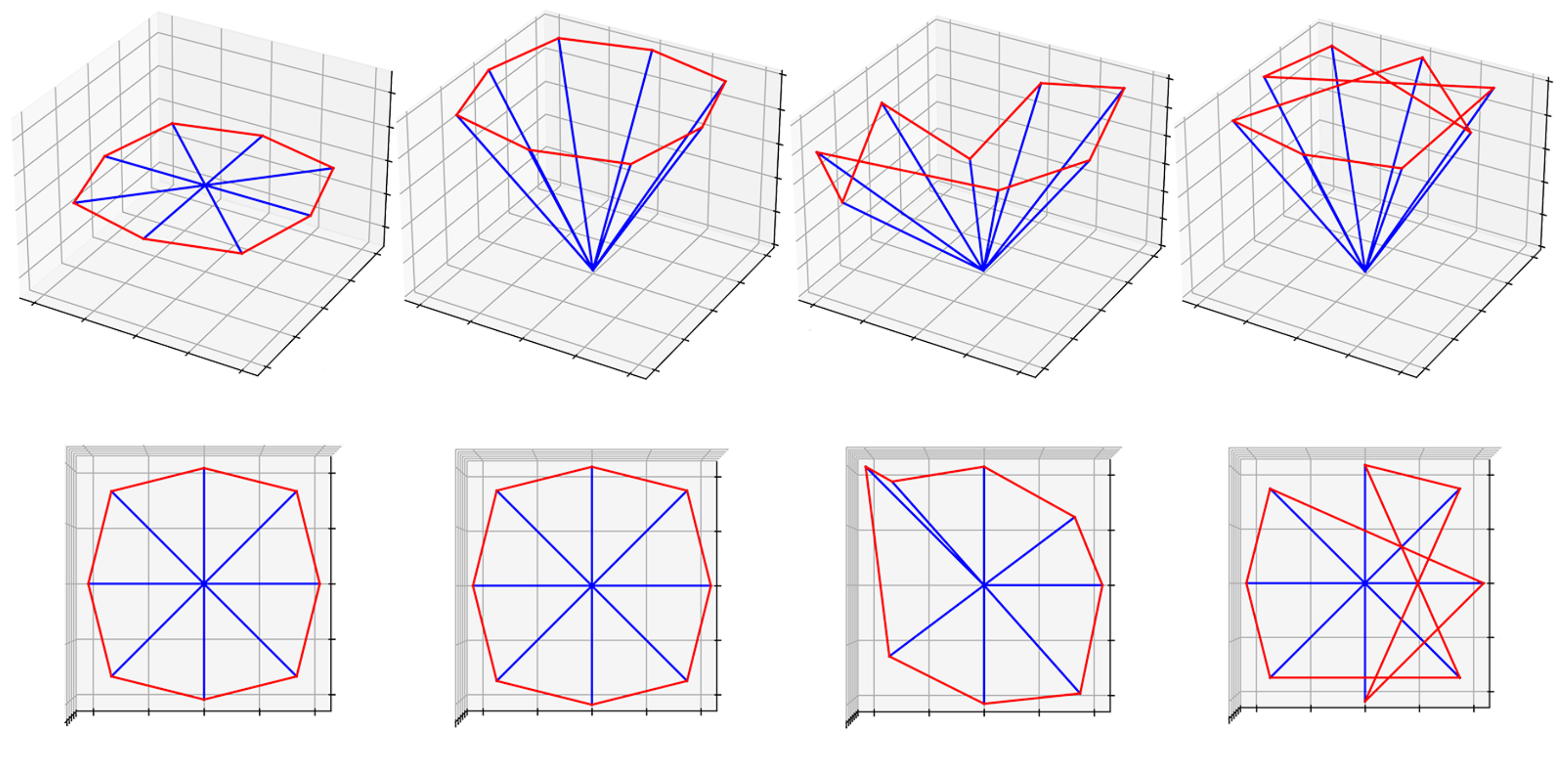}}
\end{center}
\vspace{-0.3in}
\caption{Examples of reconstructed umbrella surfaces. We present each surface with a regular view (above) and a top view (below). From left to right, we show two surfaces reconstructed from homogeneous neighbors, one from heterogeneous neighbors, and one reconstructed without sorting.}
\label{fig:umbrella}
\vspace{-0.2in}
\end{figure}

Denote the number of neighbors as $K$, the centroids and triangular features of the neighbor triangles as $\mathbf{X}_i'=\left\{\mathbf{x}_{i1}', \ldots, \mathbf{x}_{iK}'\right\} \subseteq \mathbb{R}^{K \times 3}$ and $\mathbf{T}_i=\left\{\mathbf{t}_{i1}, \ldots, \mathbf{t}_{iK}\right\} \subseteq \mathbb{R}^{K \times 4}$. In \cite{foorginejad2014umbrella}, the unsigned scalar of umbrella curvature is defined as:
\begin{equation}
u_i=\sum_{j}^{K} n_{ij} = \sum_{j}^{K} \left|\frac{\mathbf{x}_{ij}'}{\left|\mathbf{x}_{ij}'\right|} \cdot \mathbf{n}_{i}\right|,
\end{equation}
where $\mathbf{n}_i$ is the given normal vector of the $i$-th point. However, $\mathbf{n}_i$ is commonly unknown in the point set $\mathbf{X}$. This makes umbrella curvature unpractical in the real scenes. Furthermore, we argue that a scalar curvature cannot fully express the local geometry. In this case, we propose Umbrella RepSurf to express the local geometry without any given normals. Moreover, different from umbrella curvature which is defined based on homogeneous neighbors, our Umbrella RepSurf can handle heterogeneous neighbors for its position sensitivity. An illustration is shown in Fig.~\ref{fig:umbrella}. The Umbrella RepSurf $\mathbf{u}_i$ of point $\mathbf{x}_i$ is defined as:
\begin{equation}
\mathbf{u}_i=\mathcal{A}\left(\left\{\mathcal{T}\left([\mathbf{x}_{ij}', \mathbf{t}_{ij}]\right), \forall j \in \{1, \ldots, K\} \right\}\right),
\end{equation}
where $\mathcal{A}$ is an aggregation function (i.e., summation), $\mathcal{T}$ is a transformation function, and $\mathbf{x}_{ij}'$ is the normalized coordinate according to its centroid $\mathbf{x}_i$. To calculate $\mathbf{t}_{i\cdot}$, we construct adjacent triangles counterclockwise from 0$^{\circ}$ (x-axis) to 359$^{\circ}$ on the xy-plane. Thus, the number of triangles in a umbrella surface is exactly $K$. Note that, to keep local consistency of the normals' orientation, we compute these normals by counterclockwise cross product. (An example when reconstructing a umbrella surface unorderedly in Fig.~\ref{fig:umbrella}.) To simplify the definition of the global normals' orientation, different from Triangular RepSurf, we keep $a_{i1}$ of $\mathbf{t}_{i1}$ positive and the orientation of other normals changes accordingly. Therefore, though the orientation is consistent locally, the normals can be unoriented from a global perspective. Similar to Triangular RepSurf, we augment the normals of an umbrella surface $\mathbf{n}_{i\cdot}$ by random inverse. Instead of a predefined transformation function, we adopt a learnable function (a combination of linear functions and non-linearity) for $\mathcal{T}$. The implementation of Umbrella RepSurf is shown in Algorithm \ref{alg:umbrella}.

\subsection{Implementation}
\label{sec:implementation}

\begin{figure}
\begin{center}
\scalebox{0.48}{\includegraphics[width=\textwidth]{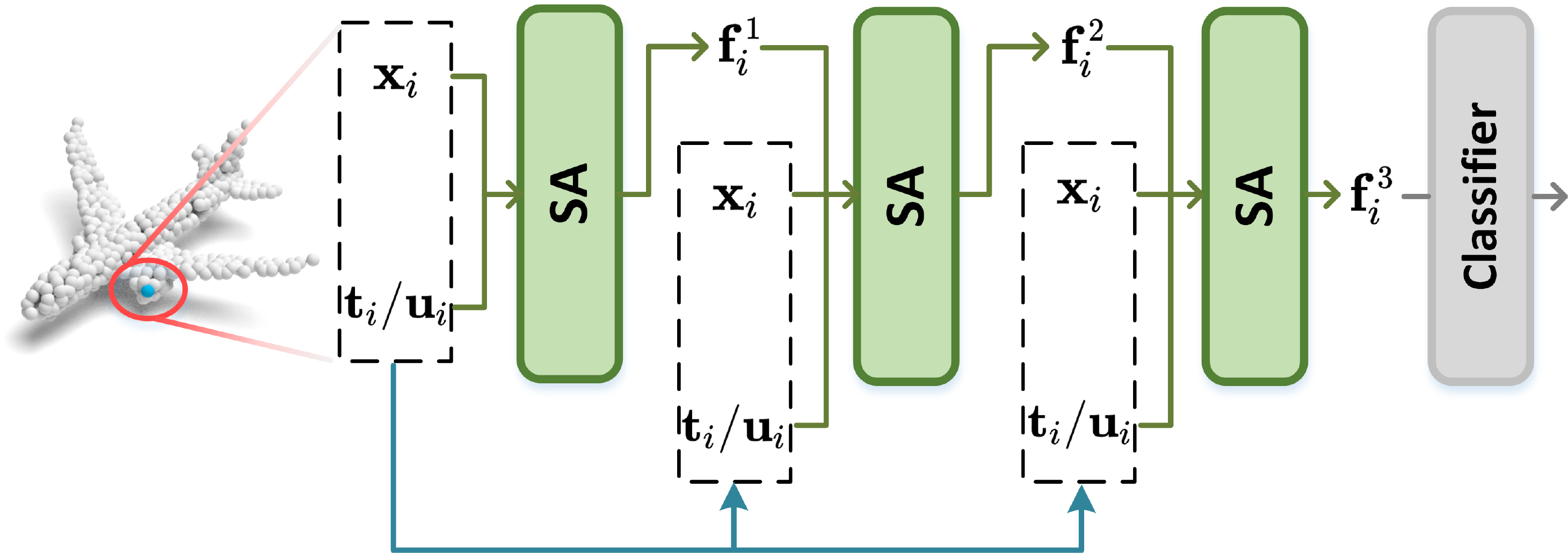}}
\end{center}
\vspace{-0.1in}
\caption{An overview of the input flow of RepSurf on PointNet++ for classification. $\mathbf{x}_i$, $\mathbf{t}_i$, $\mathbf{u}_i$ are the coordinate, Triangular RepSurf, Umbrella RepSurf of the $i$-th point of input, respectively. $\mathbf{f}^{\small 1}_i$, $\mathbf{f}^{\small 2}_i$, $\mathbf{f}^{\small 3}_i$ are the $i$-th output feature of the first, second, third set abstraction (SA), respectively.}
\label{fig:classification}
\vspace{-0.2in}
\end{figure}

We implement our RepSurf on the single-scale grouping (SSG) version of PointNet++ \cite{qi2017pointnet++} in a simple manner of concatenation. For each set abstraction, we input RepSurf along with point features. An illustration of the input flow is shown in Fig.~\ref{fig:classification}. Furthermore, we propose two designs to further improve our RepSurf.

\textbf{Polar auxiliary.} 
For simplicity, previous point-based models widely adopt Cartesian coordinates as input. However, they cannot fully express the relationships between a centroid and its neighbors. Unlike Cartesian coordinate system, the polar coordinate systems present a point coordinate by distance and angles according to the origin. The polar systems (i.e., Spherical system, Cylindrical system) can be a supplement for its distance and direction sensitivity. In this paper, we explore a practical application of the polar systems for point-based models. We take Spherical system as an example. After querying the neighbors of a point $\mathbf{x}_{i}$, we re-define the position of the $j$-th neighbor by including its spherical position:
\begin{equation}
\mathbf{x}_{ij}'=(x_{ij}',y_{ij}',z_{ij}',\rho_{ij},\theta_{ij},\phi_{ij}),
\end{equation}
where $x_{ij}'$, $y_{ij}'$, $z_{ij}'$ are the values of three dimensions of the normalized Cartesian coordinate. $\rho_{ij}=\sqrt{x_{ij}'^{2}+y_{ij}'^{2}+z_{ij}'^{2}}$, $\theta_{ij}=\arccos \frac{z_{ij}'}{\rho_{ij}}$, $\phi_{ij}=\operatorname{atan2} (y_{ij}', x_{ij}')$. For more details of the implementation on polar auxiliary, please refer to the supplementary material.

\textbf{Channel de-differentiation.} 
Inspired by \cite{yang2020cn}, we observe that different types of inputs (i.e., coordinate, normal vectors, point features) have significant differences in data distribution. In order to process different inputs equally and to train our models stably, we explore solutions for de-differentiation along the channel dimension. In this paper, we apply Post-CD (performing batch normalization after linear function) to our method. For more details of the implementation on channel de-differentiation, please refer to the supplementary material.

%%%%%%%%% Experiments

%##########################################Classification#######################################
\begin{table*}[]
\begin{center}
\scalebox{0.9}{
\addtolength{\tabcolsep}{-2.5pt}
\begin{threeparttable}
\begin{tabular}{l|c|cc|cc|cccc}
\Xhline{3\arrayrulewidth}
\multirow{2}{*}{Method}      & \multirow{2}{*}{Input}  & \multicolumn{2}{c|}{\bf ModelNet40} & \multicolumn{2}{c|}{\bf ScanObjectNN} & \multirow{2}{*}{\#Params} & \multirow{2}{*}{FLOPs} & \multirow{2}{*}{\shortstack{Train \\ Speed}} & \multirow{2}{*}{\shortstack{Infer \\ Speed}}\\ \cline{3-6}
& &  \textbf{OA} & mAcc & \textbf{OA} & mAcc  &   &   &  &  \\
\hline
PointNet \cite{qi2017pointnet}   & 1k pnts     & 89.2  & 86.0     & 68.2  & 63.4 & 3.47M & 0.45G & 1.76ms & 0.81ms  \\
DGCNN \cite{wang2019dynamic}   & 1k pnts     & 92.9  & 90.2    & 78.1  & 73.6 & 1.82M & 2.43G & - & -   \\
RS-CNN$^\ddagger$ \cite{liu2019relation}   & 1k pnts     & 93.6 & -     & - & - & 2.38M & 1.16G & - & -   \\ 
KPConv \cite{thomas2019kpconv} & $\sim$7k pnts & 92.9 & -    & - & - & 14.3M & -  & 218.7ms    & 543.7ms  \\
PointASNL \cite{yan2020pointasnl}  & ~1k pnts$^*$ & 93.2 & - & -  & -    & 10.1M   & 1.80G  & - & -  \\ 
Grid-GCN \cite{xu2020grid}   & ~1k pnts$^*$     & 93.1 & 91.3     & - & - & - & - & -   & 42.20ms  \\
PointTrans. \cite{zhao2021point}  & ~1k pnts$^*$ & 93.7 & 90.6    & - & - & -  & -  & - & -  \\  
MVTN \cite{hamdi2021mvtn} & multi-view & 93.8 & \textbf{92.0}     & 82.8 & - & 4.24M & 1.78G  & -  & -   \\
PAConv$^\ddagger$ \cite{xiang2021walk} & ~1k pnts & 93.9 & -    & - & - & 2.44M & 1.68G & - & - \\
RPNet \cite{ran2021learning} & ~1k pnts$^*$ & 94.1 & -    & - & - & 2.70M & 3.90G  & -  & -   \\
CurveNet$^\ddagger$ \cite{xiang2021walk} & ~1k pnts & 94.2 & -    & - & - & 2.14M & 0.66G & 22.04ms & 12.34ms \\
\hline

PointNet++$^\dagger$ \cite{qi2017pointnet++}  & 1k pnts  & 90.7 & 88.4   & 77.9 & 75.4 & 1.475M & 0.77G & 2.75ms  & 1.98ms \\ 

\rowcolor{RowColor} \textbf{RepSurf-T (ours)}  & 1k pnts & ~~~~~~~\textbf{94.0} \textcolor{green!40!gray}{\small $\uparrow$3.3} & ~~~~~~~91.1 \textcolor{green!40!gray}{\small $\uparrow$2.7}  & ~~~~~~~\textbf{84.1} \textcolor{green!40!gray}{\small $\uparrow$6.2} & ~~~~~~~\textbf{81.2} \textcolor{green!40!gray}{\small $\uparrow$5.8} & 1.479M & 0.79G & 3.33ms  & 2.47ms  \\ 
\rowcolor{RowColor} \textbf{RepSurf-T$^\ddagger$ (ours)}  & 1k pnts & ~~~~~~~\textbf{94.2} \textcolor{green!40!gray}{\small $\uparrow$3.5} & ~~~~~~~91.3 \textcolor{green!40!gray}{\small $\uparrow$2.9}  & ~~~~~~~\textbf{84.3} \textcolor{green!40!gray}{\small $\uparrow$6.4} & ~~~~~~~\textbf{81.6} \textcolor{green!40!gray}{\small $\uparrow$6.2} & 1.479M & 0.79G & 3.33ms  & 2.47ms  \\ 
\rowcolor{RowColor} \textbf{RepSurf-U (ours)} & 1k pnts  & ~~~~~~~\textbf{94.4} \textcolor{green!40!gray}{\small $\uparrow$3.7} & ~~~~~~~91.4 \textcolor{green!40!gray}{\small $\uparrow$3.0}   & ~~~~~~~\textbf{84.3} \textcolor{green!40!gray}{\small $\uparrow$6.4} & ~~~~~~~\textbf{81.3} \textcolor{green!40!gray}{\small $\uparrow$5.9}  & 1.483M & 0.81G & 4.08ms  & 3.10ms \\ 
\rowcolor{RowColor} \textbf{RepSurf-U$^\ddagger$ (ours)} & 1k pnts  & ~~~~~~~\textbf{94.7} \textcolor{green!40!gray}{\small $\uparrow$4.0} & ~~~~~~~91.7 \textcolor{green!40!gray}{\small $\uparrow$3.3}  & ~~~~~~~\textbf{84.6} \textcolor{green!40!gray}{\small $\uparrow$6.7} & ~~~~~~~\textbf{81.9} \textcolor{green!40!gray}{\small $\uparrow$6.5} & 1.483M & 0.81G & 4.08ms &  3.10ms  \\ \hline
\rowcolor{RowColor} \textbf{RepSurf-U$^{\ddagger\circ}$ (ours)} & 1k pnts  & - & -  & \textbf{86.0} & \textbf{83.1} & 6.806M & 2.43G & - &  -  \\ 
\Xhline{3\arrayrulewidth}
\end{tabular}
\begin{tablenotes}
\item {\small \hspace{-0.15in} $\dagger$: single-scale grouping (SSG) version, $\ddagger$: multi-scale inference from \cite{liu2019relation}, $*$: w/ normal vector, $\circ$: PointNet++ (SSG) with double channels and deeper networks.}
\end{tablenotes}
\end{threeparttable}}
\end{center}
\vspace{-0.2in}
\caption{Performance of classification on ModelNet40 and ScanObjectNN. We evaluate different methods in terms of \textbf{overall accuracy} (OA, \%), mean per-class accuracy (mAcc, \%), number of parameters (\#Params), FLOPs, training speed (duration per input sample), and inference speed (duration per input sample). We consider \textbf{OA} the principle evaluation metric. \textbf{Bold} means the result outperforms prior state-of-the-art method on corresponding dataset. \textcolor{green!40!gray}{Green} means an improvement from our RepSurf compared with the original model. We test the speed of all methods with one NVIDIA Tesla V100 GPU and four cores of Intel Xeon @2.50GHz CPU. The batch size is set to 16.}
\label{tab:classification}
\end{table*}
%##################################################################################################

%##########################################Semantic Segmentation#######################################
\begin{table*}
\begin{center}
\scalebox{0.9}{
\addtolength{\tabcolsep}{-5pt}
\begin{threeparttable}
\begin{tabular}{l|ccc|ccc|c|cc}
\Xhline{3\arrayrulewidth}
\multirow{2}{*}{Method} & \multicolumn{3}{c|}{\bf S3DIS 6-fold} & \multicolumn{3}{c|}{\bf S3DIS Area-5}  & {\bf ScanNet} & \multirow{2}{*}{\#Params}  & \multirow{2}{*}{FLOPs} \\ \cline{2-8}
& mIoU & mAcc & OA & mIoU & mAcc & OA & mIoU & & \\
\hline
PointNet \cite{qi2017pointnet}     & 47.6  & 66.2 & 78.5  & 41.1 & 48.9 & -    & -  &  1.7M & 4.1G     \\
PointWeb     \cite{zhao2019pointweb}   & 66.7  & 76.2 & 87.3   &  60.2  & 66.6 & 86.9  & -           & -      & -         \\
KPConv \cite{thomas2019kpconv} & 70.6 & 79.1 & - & 67.1 & 72.8 & -     & 68.4 & 14.9M   & -    \\
PointASNL \cite{yan2020pointasnl} & 68.7 & 79.0 & 88.8  & 62.6 & 68.5 & 87.7   & 63.0  & 22.4M  & 19.1G   \\ 
PAConv \cite{xiang2021walk} & 69.3 & 78.6 & - & 66.5 & 73.0 & -  & - & - & 1.3G \\
RPNet \cite{ran2021learning} & 70.8 & - & -  & -  & - & -   & 68.2 & 2.4M  & 5.1G      \\
PointTrans. \cite{zhao2021point}  & 73.5  & 81.9 & 90.2 & \textbf{70.4} & \textbf{76.5} & \textbf{90.8} & -  & 4.9M  & 2.8G   \\  
\hline
PointNet++$^\dagger$     \cite{qi2017pointnet++}   & 59.9   & 66.1 & 87.5   & 56.0   & 61.2 & 86.4  & - &  0.969M   &  1.00G          \\

\rowcolor{RowColor} \textbf{RepSurf-U (ours)}  
& ~~~~~~~~\textbf{74.3}\textcolor{green!40!gray}{\small $\uparrow$14.4}  
& ~~~~~~~~\textbf{82.6}\textcolor{green!40!gray}{\small $\uparrow$16.5} 
& ~~~~~~\textbf{90.8}\textcolor{green!40!gray}{\small $\uparrow$3.3} 
& ~~~~~~~~68.9\textcolor{green!40!gray}{\small $\uparrow$12.9}    
& ~~~~~~~~76.0\textcolor{green!40!gray}{\small $\uparrow$14.8}              
& ~~~~~~90.2\textcolor{green!40!gray}{\small $\uparrow$3.8}      

& \textbf{70.0}
&  0.976M   &  1.04G \\ 

\Xhline{3\arrayrulewidth}
\end{tabular}
\begin{tablenotes}
\item {\small \hspace{-0.15in} $\dagger$: single-scale grouping (SSG) version, $*$: w/ normal vector.}
\end{tablenotes}
\end{threeparttable}
}
\end{center}
\vspace{-0.2in}
\caption{Performance of semantic segmentation on S3DIS (evaluation by 6-fold or on Area 5) and ScanNet~V2. We evaluate different methods in terms of mean per-class IoU (mIoU, \%), mean per-class accuracy (mAcc, \%), overall point accuracy (OA, \%), number of parameters (\#Params), and FLOPs. \textbf{Bold} means the result outperforms prior state-of-the-art method on corresponding dataset. \textcolor{green!40!gray}{Green} means an improvement from our RepSurf compared with the previous reported results of the original model.}
\label{tab:semantic}
\end{table*}
%##################################################################################################

\section{Experiments}

We evaluate both of our Triangular RepSurf and Umbrella RepSurf on three main tasks: classification, segmentation, and detection. Furthermore, we conduct ablation studies to assess the effectiveness of our designed modules. Please refer to the supplementary material for more experimental details.

\subsection{Classification}

3D object classification is a basic task to prove the effectiveness of methods. We perform experiments on ModelNet40 \cite{wu20153d}, a human-made object dataset, and ScanObjectNN \cite{uy2019revisiting}, a dataset retrieved from the real scenes.

\textbf{Human-made Object Classification.}~~ModelNet40 \cite{wu20153d} contains 9843 training models and 2468 test models, divided into 40 categories. In Tab.~\ref{tab:classification}, we compare our Triangular RepSurf (RepSurf-T) and Umbrella RepSurf (RepSurf-U) with prior methods. Equipped with RepSurf-T and RepSurf-U, the performance of PointNet++ (SSG version) is considerably boosted by 3.7\% and 4.1\%. For a fair comparison with other methods \cite{xiang2021walk, liu2019relation, xu2021paconv}, we apply the strategy of multi-scale inference from \cite{liu2019relation} for further improvement. Though the results on ModelNet40 tend to be saturated, our RepSurf-U achieves 94.7\%, surpassing CurveNet \cite{xiang2021walk} by a large margin of 0.5\%. In addition, RepSurf-U is 5.4$\times$ and 4.0$\times$ faster than CurveNet in terms of training and inference speed, respectively.

\textbf{Real-world Object Classification.}~~For the saturation of ModelNet40, we further verify our RepSurf on the hardest variant (PB\_T50\_RS variant) of ScanObjectNN \cite{wu20153d}, a more challenging dataset considering occlusion and background. It is composed of 2902 point clouds categorized into 15 classes. In Tab.~\ref{tab:classification}, our RepSurf-T and RepSurf-U achieve 84.3\% and 84.6\%, outperforming prior state-of-the-art MVTN \cite{hamdi2021mvtn} by 1.5\% and 1.8\%, with around 1.8$\times$ fewer parameters and 1.2$\times$ fewer FLOPs.

%###########################################Detection#################################
\begin{table*}[]
\begin{center}
\scalebox{0.9}{
\addtolength{\tabcolsep}{2.5pt}
\begin{threeparttable}
\begin{tabular}{l|c|cc|cc|cc}
\Xhline{3\arrayrulewidth}
\multirow{2}{*}{Method} & \multirow{2}{*}{Backbone} & \multicolumn{2}{c|}{\bf ScanNetV2} & \multicolumn{2}{c|}{\bf SUN RGB-D } & \multirow{2}{*}{\#Params} & \multirow{2}{*}{\shortstack{Infer \\ Speed}}  \\ \cline{3-6}
&  &  \text{mAP@0.25} & \text{mAP@0.5} & \text{mAP@0.25} & \text{mAP@0.5} & & \\
\hline 
VoteNet \cite{qi2019deep} & PointNet++ & 62.9 & 39.9 & 59.1 & 35.8 & -  & - \\
ImVoteNet \cite{qi2020imvotenet} & PointNet++ & - & - & ~63.4$^*$ & - & - & - \\
H3DNet \cite{zhang2020h3dnet} & PointNet++ & 64.4 & 43.4 & - & - & - & -  \\ 
H3DNet \cite{zhang2020h3dnet} & 4$\times$PointNet++ & 67.2 & 48.1   & 60.1 & 39.0 & - & 266ms \\ 
3DETR \cite{misra2021end} & Transformer & 65.0 & 47.0 & 59.1 & 32.7 & -  & - \\
BRNet \cite{cheng2021back} & PointNet++ & 66.1 & 50.9 & 61.1 & 43.7 & - & - \\
\hline

GroupFree$^{6, 256}$ & PointNet++ & 67.3 & 48.9 & 63.0  & 45.2  & 11.49M & 149ms \\
\rowcolor{RowColor} GroupFree$^{6, 256}$ & \textbf{RepSurf-T} 
&  ~~~~~~~\textbf{68.4} \textcolor{green!40!gray}{\small $\uparrow$1.1} 
&  ~~~~~~~50.3 \textcolor{green!40!gray}{\small $\uparrow$0.4} 
&  ~~~~~~~\textbf{63.9} \textcolor{green!40!gray}{\small $\uparrow$0.9}
&  ~~~~~~~\textbf{45.6} \textcolor{green!40!gray}{\small $\uparrow$0.4}
& 11.50M & 149ms \\

\rowcolor{RowColor} GroupFree$^{6, 256}$ & \textbf{RepSurf-U} 
& ~~~~~~~\textbf{68.8} \textcolor{green!40!gray}{\small $\uparrow$1.5} 
& ~~~~~~~50.5 \textcolor{green!40!gray}{\small $\uparrow$0.6} 
& ~~~~~~~\textbf{64.3} \textcolor{green!40!gray}{\small $\uparrow$1.3} 
& ~~~~~~~\textbf{45.9} \textcolor{green!40!gray}{\small $\uparrow$0.7} 
& 11.50M & 150ms \\
\hline

GroupFree$^{12, 512}$ & PointNet++$^2$ & 69.1 & 52.8  & - & -  & 23.60M & 193ms \\
\rowcolor{RowColor} GroupFree$^{12, 512}$ & \textbf{RepSurf-T}$^2$ 
& ~~~~~~~\textbf{70.4} \textcolor{green!40!gray}{\small $\uparrow$1.3} 
& ~~~~~~~\textbf{54.6} \textcolor{green!40!gray}{\small $\uparrow$1.8} 
& \textbf{64.2}  
& \textbf{47.1} 
& 23.60M & 194ms \\

\rowcolor{RowColor} GroupFree$^{12, 512}$ & \textbf{RepSurf-U}$^2$ 
& ~~~~~~~\textbf{71.2} \textcolor{green!40!gray}{\small $\uparrow$2.1} 
& ~~~~~~~\textbf{54.8} \textcolor{green!40!gray}{\small $\uparrow$2.0}  
& \textbf{64.9} 
& \textbf{47.7} 
& 23.61M & 195ms \\
\Xhline{3\arrayrulewidth}
\end{tabular}
\begin{tablenotes}
\item {\small \hspace{-0.15in} $*$: w/ RGB as input, Model$^2$: Model with doubled channels for each MLP, 4$\times$PointNet++: four individual PointNet++ (SSG) in \cite{zhang2020h3dnet}, GroupFree$^{a, b}$: GroupFree model \cite{liu2021group} with a $a$-layer decoder and $b$ object candidates.}
\end{tablenotes}
\end{threeparttable}
}
\end{center}
\vspace{-0.2in}
\caption{Performance of object detection on ScanNet~V2 and SUN RGB-D. We evaluate different methods in terms of mAP@0.25, mAP@0.5, number of parameters (\#Params), and inference speed (duration per input sample). \textbf{Bold} means the result outperforms prior state-of-the-art method on corresponding dataset. \textcolor{green!40!gray}{Green} means an improvement from our RepSurf compared with the original model. We test the speed of all methods with one NVIDIA Titan-XP GPU and four cores of Intel Xeon @2.50GHz CPU.}
\label{tab:detection}
\end{table*}
%##################################################################################################

%------------------------------------------------------------------------
\subsection{Segmentation}

Scene segmentation can be more challenging due to outliers and noise. We evaluate our RepSurf on two large-scale scene datasets, S3DIS \cite{armeni20163d} and ScanNet V2 \cite{dai2017scannet}.

\textbf{Semantic Segmentation on S3DIS.}~~S3DIS \cite{armeni20163d} contains 271 scenes from 6 indoor areas. Each point is categorized into 13 types of semantic labels. In Tab.~\ref{tab:semantic}, we evaluate our RepSurf on S3DIS by 6-fold and on Area-5. Our RepSurf-U significantly improves PointNet++ by 14.4\% mIoU and 12.9\% mIoU on S3DIS 6-fold and S3DIS Area-5, respectively. Furthermore, our RepSur-U outperforms previous state-of-the-art, Point Transformer \cite{zhao2021point} by 0.8\% mIoU on S3DIS 6-fold, and achieves comparable performance on S3DIS Area-5 as well. Simultaneously, our RepSurf-U has 4.0$\times$ fewer parameters and 1.7$\times$ fewer FLOPs with a comparison of Point Transformer.

\textbf{Semantic Segmentation on ScanNet.}~~ScanNet V2 \cite{dai2017scannet} consists of 1513 indoor training point clouds and 100 test point clouds. It marks each point with 21 categories. In Tab.~\ref{tab:semantic}, the performance of RepSurf-U exceeds prior state-of-the-art KPConv \cite{thomas2019kpconv} by 1.6\%. Moreover, our method has 14.3$\times$ fewer parameters compared with KPConv.

%------------------------------------------------------------------------
\subsection{Detection}

3D detection can further prove the superiority of our method at the application level. We conduct experiments on two widely adopted 3D object detection datasets: ScanNet V2~\cite{dai2017scannet} and SUN RGB-D~\cite{song2015sun}. We adopt a powerful method \cite{liu2021group} for pipeline and replace the backbone with our RepSurf to perform all experiments on this task. Our experiments are mainly based on the codebase$^1$\footnote{$^1$\url{https://github.com/zeliu98/Group-Free-3D}} of \cite{liu2021group} as well.

\textbf{Detection on ScanNet.}~~ScanNet V2~\cite{dai2017scannet} can be adopted for 3D detection as well, consisting of 1513 indoor scenes and 18 object classes. We follow the standard evaluation protocol in \cite{qi2019deep} by utilizing mean Average Precision under the thresholds of 0.25 (mAP@0.25) and 0.5 (mAP@0.5), without considering the orientation of bounding boxes. As shown in Tab.~\ref{tab:detection}, with almost no increase in computational cost ($\sim$0.01M parameters and $\sim$1ms inference speed), our RepSur-U boosts the performance of previous state-of-the-art \cite{liu2021group} by 2.1\% mAP@0.25 and 2.0\% mAP@0.25.

\textbf{Detection on SUN RGB-D.}~~SUN RGB-D~\cite{song2015sun} is a single-view RGB-D dataset for 3D scene analysis, including around 5K indoor RGB and depth images. Following \cite{qi2019deep}, we adopt mean Average Precision on 10 most common categories for evaluation. In Tab.~\ref{tab:detection}, RepSurf-U improves GroupFree$^{6, 256}$ (\cite{liu2021group} with a 6-layer encoder and 256 object candidates) by 1.3\% mAP@0.25 and 0.7\% mAP@0.5. Without RGB as input, GroupFree$^{12, 512}$ equipped with RepSurf-U even outperforms prior state-of-the-art ImVoteNet \cite{qi2020imvotenet} by 1.5\% mAP@0.25.

%----------------------------------------

\subsection{Ablation study}
\label{sec:ablation}

%----------------------------------------
\begin{table}[]
\begin{center}
\scalebox{1}{
\addtolength{\tabcolsep}{0pt}
\begin{tabular}{cccc|c}
\Xhline{3\arrayrulewidth}
type & $\mathcal{X}$-computed & w/ $p_i$ & w/ inverse & acc. \\ 
\hline
given & - & \xmark  & \xmark & 94.08 \\
given & - & \xmark  & \cmark & 93.39  \\ 
given & - & \cmark  & \xmark & 93.95  \\ 
\hline
triangular & pre & \cmark & \xmark & 93.57  \\
triangular & post &  \cmark & \xmark & 93.62  \\
\rowcolor{RowColor} triangular & post &  \cmark & \cmark  & \textbf{94.02} \\
\hline
umbrella & pre & \cmark & \xmark & 93.06  \\
umbrella & post &  \cmark  & \xmark & 93.90 \\
\rowcolor{RowColor} umbrella & post &  \cmark & \cmark & \textbf{94.46} \\
\Xhline{3\arrayrulewidth}
\end{tabular}}
\end{center}
\vspace{-0.2in}
\caption{Ablation study on the types of RepSurf. (given: normal vectors given from the dataset, triangular: Triangular RepSurf, umbrella: Umbrella RepSurf, $\mathcal{X}$-computed: computing RepSurf before (pre-computed) or after (post-computed) sampling, w/ $p_i$: with surface position $p_i$ input, w/ inverse: augmenting RepSurf by random inverse, acc.: overall accuracy)}
\label{tab:ablate_type}
\vspace{-0.2in}
\end{table}

We ablate some vital designs of our method on ModelNet40 for an insightful exploration.

\textbf{Types of RepSurf.}~~Shown in Tab.~\ref{tab:ablate_type}, we compare different types of input (given normals, RepSurf-T, RepSurf-U). We further discuss on when to compute RepSurf. Regularly, we obtain the input point clouds after a process of sampling (i.e., 10000 $\rightarrow$ 1024 points). Before this process (pre-computed), we will derive RepSurf from high-resolution point clouds, which means RepSurf approximates the corresponding tangent. However, empirical results show that post-computed works better than pre-computed. We additionally test on the designs of surface position and random inverse, both of which slightly improve RepSurf.

%----------------------------------------

\textbf{Design of RepSurf block.}~~Shown in Tab.~\ref{tab:ablate_block}, we explore the design of Umbrella RepSurf in terms of input, transformation function $\mathcal{T}$, and aggregation function $\mathcal{A}$. Empirically, a combination of normal vector, surface position, normalized coordinate and the corresponding polar coordinates outperforms other combinations. Furthermore, prohibition of batch norm, usage of bias for the first layer, sum-pooling, and three-layer MLP perform better than other options.

\begin{table}[]
\begin{center}
\scalebox{1}{
\begin{threeparttable}
\addtolength{\tabcolsep}{-2.pt}
\begin{tabular}{cccccc|c}
\Xhline{3\arrayrulewidth}
input & \#channels & BN & bias & $\mathcal{A}$ & \#layers & acc. \\ 
\hline
N & 3 & \xmark & \xmark &  sum & 1 & 93.17 \\
N+P & 4 & \xmark & \xmark & sum & 1 & 93.24 \\
N+C & 6 & \xmark & \xmark & sum & 1 &  93.18 \\
N+P+C & 7 & \xmark & \xmark & sum & 1 & 93.38  \\
N+P+CP & 10 & \xmark & \xmark & sum & 1 & 93.45 \\ 
\hline
N+P+CP & 10 & \xmark & \cmark & sum & 1 & 93.86 \\
N+P+CP & 10 & \xmark & \cmark & sum & 2 &  93.94 \\
\hline
N+P+CP & 10 & \xmark & \cmark & max & 3 & 94.04 \\
N+P+CP & 10 & \xmark & \cmark & mean & 3 & 94.37 \\
N+P+CP & 10 & \cmark & \cmark & sum & 3 & 94.06 \\ 
\rowcolor{RowColor} N+P+CP & 10 & \xmark & \cmark & sum & 3 & \textbf{94.46} \\ 
\Xhline{3\arrayrulewidth}
\end{tabular}
\end{threeparttable}}
\end{center}
\vspace{-0.2in}
\caption{Ablation study on the design of Umbrella RepSurf block. (N: normal vector $(a_i, b_i, c_i)$, P: surface position $p_i$, C: centroid position $(x_{ij}',y_{ij}',z_{ij}')$, CP: centroid position $(x_{ij}',y_{ij}',z_{ij}')$ with polar auxiliary $(\rho_{ij},\theta_{ij},\phi_{ij})$), \#channels: number of input channels, BN: applying batch normalization, bias: applying learnable bias in the first layer, $\mathcal{A}$: aggregation function, \#layers: number of MLP layers for mapping, acc.: overall accuracy)}
\label{tab:ablate_block}
\vspace{-0.2in}
\end{table}

\textbf{Group size.}~~We explore the group size of Umbrella RepSurf in terms of both accuracy and speed (ms per sample):
\\[4pt]
\centerline{
\addtolength{\tabcolsep}{-3pt}
\begin{tabular}{l|ccccccc}
PN2 & $k$=2 & $k$=4 & $k$=6 & {\bf ${\boldsymbol k}$=8} & $k$=10 & $k$=12 & $k$=16  \\ 
\Xhline{3\arrayrulewidth}
acc. & 93.53 & 93.63 & 94.36 & \textbf{94.46} & 94.32 & 94.20 & 94.32 \\
time & 0.50 & 0.46 & 0.49 & 0.48 & 0.58 & 0.51 & 0.78
\end{tabular}} 
\\[4pt]
We test the speed of Umbrella RepSurf block only. When $k$=2, Umbrella RepSurf will degenerate to a learnable version of Triangular RepSurf. There is almost no difference in speed when $k$ is in the range of $[2, 12]$. For a trade-off between performance and speed, we consider $k$=8 an ideal choice. Furthermore, when we study on larger group sizes (i.e., 24), a vanishing gradient problem exists. We argue that larger umbrella surfaces may become more indistinguishable and lead to the problem, but this is still an open issue.

\textbf{Polar auxiliary.}~~We study on the design of our polar auxiliary in different versions: 
\\[4pt]
\centerline{
\addtolength{\tabcolsep}{-3pt}
\begin{tabular}{l|cccc}
PN2 & w/o aux.  & w/ $\rho$ & w/ cylinder & w/ sphere \\ 
\Xhline{3\arrayrulewidth}
acc. & 93.97 & 94.12 \textcolor{green!40!gray}{\small $\uparrow$0.15} & 93.89 \textcolor{red!70!gray}{\small $\downarrow$0.08} & \textbf{94.46}  \textcolor{green!40!gray}{\small $\uparrow$0.49}              
\end{tabular}} 
\\[4pt]
Here $\rho$, a part of spherical polar auxiliary, means the distance between a centroid and its neighbors. We discuss that Spherical system can better express the geometric relations between the centroids and their neighbors, an auxiliary of Cartesian system. Empirical results verify this hypothesis.

\begin{figure}
\begin{center}
\scalebox{0.35}{
    \includegraphics[width=\textwidth]{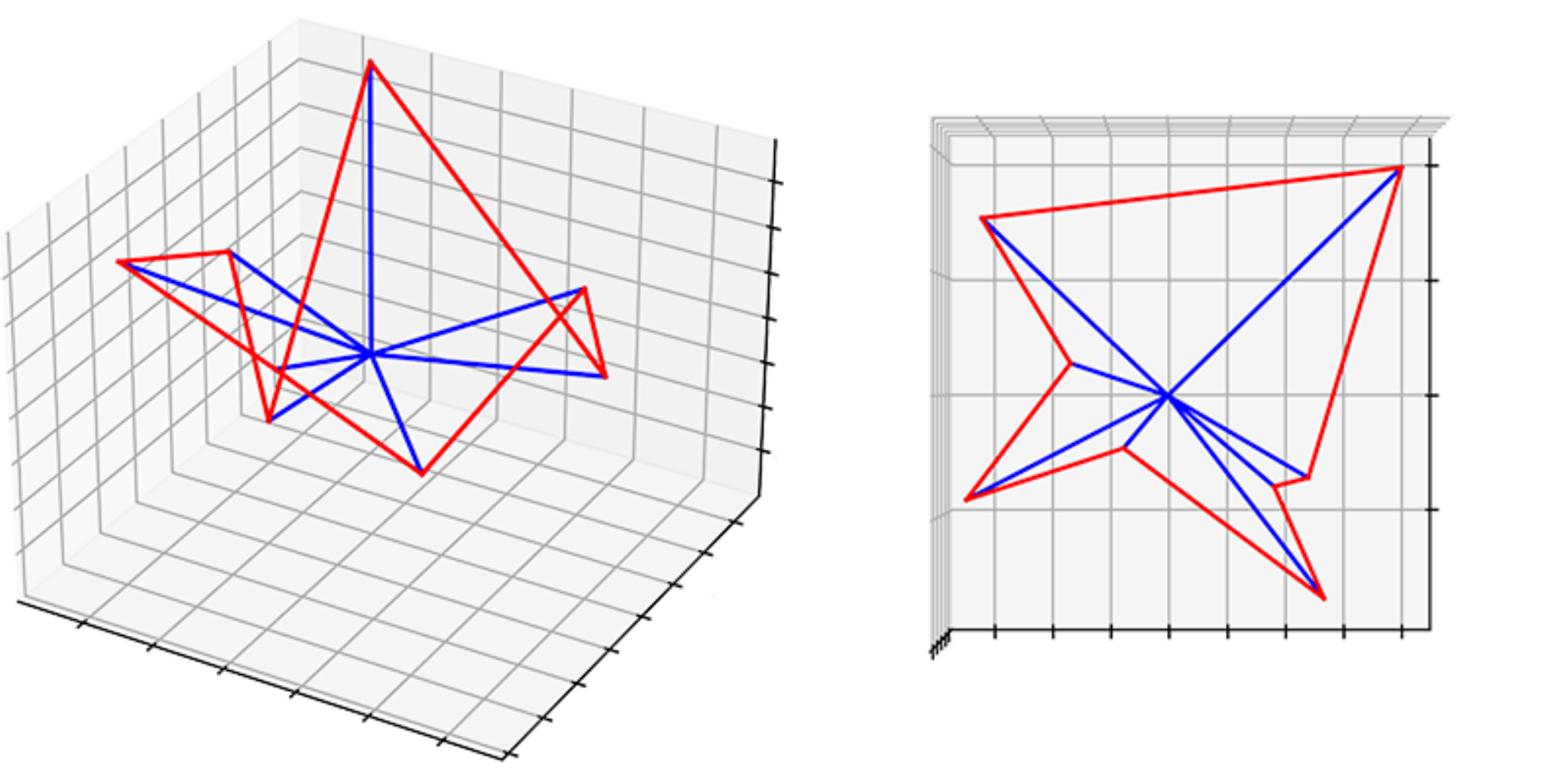}
}
\end{center}
\vspace{-0.2in}
\caption{Bad case of a reconstructed umbrella surface when the neighbors are extremely messy.}
\label{fig:bad}
\vspace{-0.2in}
\end{figure}

\textbf{Channel de-differentiation.}~~We test the design of channel de-differentiation (CD) on three versions of PointNet++, including the original (vanilla), Triangular RepSurf (triangular), and Umbrella RepSurf (umbrella):
\\[4pt]
\centerline{
\addtolength{\tabcolsep}{1pt}
\begin{tabular}{l|ccc}
PN2 & none  & Pre-CD & Post-CD  \\ 
\Xhline{3\arrayrulewidth}
vanilla 
& 93.15
& 92.70 \textcolor{red!70!gray}{\small $\downarrow$0.36}
& \textbf{94.08} \textcolor{green!40!gray}{\small $\uparrow$0.93} \\
triangular 
& 93.22
& 92.49 \textcolor{red!70!gray}{\small $\downarrow$0.73} 
& \textbf{94.02} \textcolor{green!40!gray}{\small $\uparrow$0.80} \\
umbrella 
& 93.50
& 92.63 \textcolor{red!70!gray}{\small $\downarrow$0.87}
& \textbf{94.46} \textcolor{green!40!gray}{\small $\uparrow$0.96} \\ 
\end{tabular}}
\\[4pt]
Here Pre-CD means that batch normalization performs before linear function, and Post-CD is the opposite. We argue Post-CD performs better than Pre-CD, since Pre-CD may blur the original semantics of external input (i.e., coordinates, RepSurf features).

%%%%%%%%% Discussion
\section{Discussion}
\textbf{Limitation.}~~Though simple and effective, RepSurf may suffer from noises while surface reconstruction due to the noise-sensitive algorithm kNN. Furthermore, we argue that Umbrella RepSurf may be vulnerable to extremely messy points. Thus, when we query more neighbors of a point, in general the distribution of its neighbors would become messy and results in a distorted surface. An example of bad case is shown in Fig.~\ref{fig:bad}.

\textbf{Conclusion.}~~We present two variants of RepSurf, Triangular and Umbrella RepSurf, to explore the surface representation on point clouds. We evaluate our simple baseline on various tasks, including shape classification, scene segmentation and detection. The evaluation results show its astonishing efficiency and performance, superior to the previous state-of-the-art on different benchmarks.

We hope our work can inspire the community and evoke the rethinking on the explicit representation of point clouds. We believe that RepSurf deserves further exploration for different fields (i.e., autonomous driving) or on larger-scale point clouds, since RepSurf is eligible to handle numerous background points in the real scenes. RepSurf may also be helpful for point cloud sampling by its ability on geometry sensitivity. It would be worthy of solving the above limitations of RepSurf as well.

{\small
\bibliographystyle{ieee_fullname}
\bibliography{main}
}

\clearpage

\appendix

\noindent{\Large \bf Appendix}

\section{Preliminaries: Taylor Series for 2D curves}
\label{sec:preliminaries}

Talyor series \cite{taylor1717methodus} on the point $(a, f(a))$ of curve $f(\cdot)$ presents as follows:
\begin{equation}
f(a)+\frac{f^{\prime}(a)}{1 !}(x-a)+\frac{f^{\prime \prime}(a)}{2 !}(x-a)^{2}+\frac{f^{\prime \prime \prime}(a)}{3 !}(x-a)^{3}+\cdots,
\end{equation}
which can be simplified as:
\begin{equation}
\sum_{n=0}^{\infty} \frac{f^{(n)}(a)}{n !}(x-a)^{n},
\end{equation}
where $f^{(n)}(a)$ is the $n$-th derivative of the curve $f(\cdot)$ at the point $(a, f(a))$. 

We present the assumption that the formulation of Taylor Series can depict the local curve. Based on this assumption, we further develop an extension to 3D space.

\section{Preliminaries: Two-Variate Taylor Series for 3D surfaces}
\label{sec:preliminaries}

Taylor Series depending on two variables can be defined as:
\begin{equation}
g(a, b)+\frac{1}{1 !}(x-a, y-b) \cdot\left(\begin{array}{l}
\frac{\partial g}{\partial x}(a, b) \\
\frac{\partial g}{\partial y}(a, b)
\end{array}\right) + \cdots,
\end{equation}
where $\frac{\partial g}{\partial x}$ and $\frac{\partial g}{\partial y}$ are the partial derivatives. This formulation presents two-variant taylor series on point $(a, b, g(a,b))$ of surface $g(\cdot, \cdot)$. 

This formulation reveals the basis of RepSurf. To simply the calculation, we consider the terms of the first and second partial derivatives. Triangular RepSurf can be an instantiation. 

\section{Details of Polar Auxiliary}
\label{sec:polar}
We present two types of polar auxiliary, spherical and cylindrical ones based on Spherical Polar System and Cylindrical Polar System, respectively. 

For a given point $(x, y, z)$, spherical polar auxiliary provides the corresponding polar coordinate $(\rho_s,\theta_s,\phi_s)$, where $\rho_s=\sqrt{x^{2}+y^{2}+z^{2}}\in [0, +\infty)$, $\theta_s=\arccos \frac{z}{\rho}\in [0, \pi]$, $\phi_s=\operatorname{atan2} (y, x)\in [0, 2\pi)$. For stable training, we normalize the polar coordinate by $\theta_s$ divided by $\pi$ and $\phi_s$ divided by $2\pi$. Though $\rho_s$ has no upper bound in theory, $\rho_s$ is commonly limited within $[0, r]$, where $r$ is the radius of ball query function \cite{qi2017pointnet++}. Furthermore, to prevent the generation of NaN, we set $\theta_s$ to 0 when $\rho_s$ is 0. The pseudo-code of spherical polar auxiliary is presented in Algorithm~\ref{alg:spherical}.

Accordingly, cylindrical polar auxiliary from $(x, y, z)$ gives the polar coordinate $(\rho_c,\theta_c,z_c)$, where $\rho_s=\sqrt{x^{2}+y^{2}}\in [0, r]$, $\phi_s=\operatorname{atan2} (y, x)\in (-\pi, \pi)$, $z_c = z\in [-r, r]$, $r$ is the given radius of ball query function \cite{qi2017pointnet++}. Similarly, we normalize $\phi_s$ and $z_c$ into the range of $[0, 1]$. We implement polar auxiliary by concatenation of the Cartesian coordinate $(x,y,z)$ and $(\rho_s,\theta_s,\phi_s)$ or $(\rho_c,\theta_c,z_c)$. The pseudo-code of cylindrical polar auxiliary in Algorithm~\ref{alg:cylindrical}.

Though extremely simple, our design of polar auxiliary is not an incremental method and can be insightful. Polar auxiliary is mainly relied upon the prerequisite that the models learn the local shapes within the queried balls. This prerequisite allows spherical polar coordinate to work with Cartesian coordinate more reasonably. We argue that a Cartesian coordinate is efficient to represent the location of a point numerically according to the origin or the centroid. However, it cannot obviously discriminate the locations of two neighbors. When the two points are very close, Cartesian coordinates show few clues to tell both. In this case, $\theta_s$ and $\phi_s$ can intuitively magnify the difference between the two points numerically. Furthermore, $\rho_s$ is an additional ingredient to express the relationship between a neighbor point and its centroid. Both empirical results and theoretical analysis prove the effectiveness of our design of polar auxiliary.

%##########################################Code of Spherical Polar Auxiliary#####################################
\begin{algorithm}[t]
\caption{Pytorch-Style Pseudocode of Spherical Polar Auxiliary}
\label{alg:spherical}
\definecolor{codeblue}{rgb}{0.25,0.5,0.7}
\vspace{-4pt}
\lstset{
  backgroundcolor=\color{white},
  basicstyle=\fontsize{7.6pt}{7.6pt}\ttfamily\selectfont,
  columns=fullflexible,
  breaklines=true,
  captionpos=b,
  commentstyle=\fontsize{10pt}{10pt}\color{codeblue},
  keywordstyle=\fontsize{10pt}{10pt},
}
\begin{lstlisting}[language=python]
# xyz: coordinates of a point set
rho = sqrt(sum(pow(xyz,2),dim=-1,keepdim=True))
rho = clamp(rho,min=0) # range: [0, inf]
theta = acos(xyz[...,2,None]/rho) # range: [0, pi]
phi = atan2(xyz[..., 1,None], xyz[..., 0,None]) # range: [-pi, pi]

# check nan
idx = rho==0
theta[idx] = 0

# normalize
theta = theta/pi # [0, 1]
phi = phi/(2*np.pi)+.5 # [0, 1]
out = torch.cat([rho,theta,phi],dim=-1)
return out
\end{lstlisting}
\vspace{-3pt}
\label{alg:pseudocode}
\end{algorithm}

%##################################################################################################

%##########################################Code of Cylindrical Polar Auxiliary#####################################
\begin{algorithm}[t]
\caption{Pytorch-Style Pseudocode of Cylindrical Polar Auxiliary}
\label{alg:cylindrical}
\definecolor{codeblue}{rgb}{0.25,0.5,0.7}
\vspace{-4pt}
\lstset{
  backgroundcolor=\color{white},
  basicstyle=\fontsize{7.6pt}{7.6pt}\ttfamily\selectfont,
  columns=fullflexible,
  breaklines=true,
  captionpos=b,
  commentstyle=\fontsize{10pt}{10pt}\color{codeblue},
  keywordstyle=\fontsize{10pt}{10pt},
}
\begin{lstlisting}[language=python]
# xyz: coordinates of a point set
rho = sqrt(sum(pow(xyz[...,:2],2),dim=-1,keepdim=True))
rho = clamp(rho,0,1) # range: [0, 1]
phi = atan2(xyz[...,1,None], xyz[...,0,None]) # range: [-pi, pi]
z = xyz[...,2,None]
z = torch.clamp(z,-1,1) # range: [-1, 1]

# normalize
phi = phi/(2*pi)+.5
z = (z+1.)/2.
out = torch.cat([rho,phi,z],dim=-1)
return out
\end{lstlisting}
\vspace{-3pt}
\label{alg:pseudocode}
\end{algorithm}

%##################################################################################################

\begin{figure*}
\begin{center}
    \includegraphics[width=0.9\textwidth]{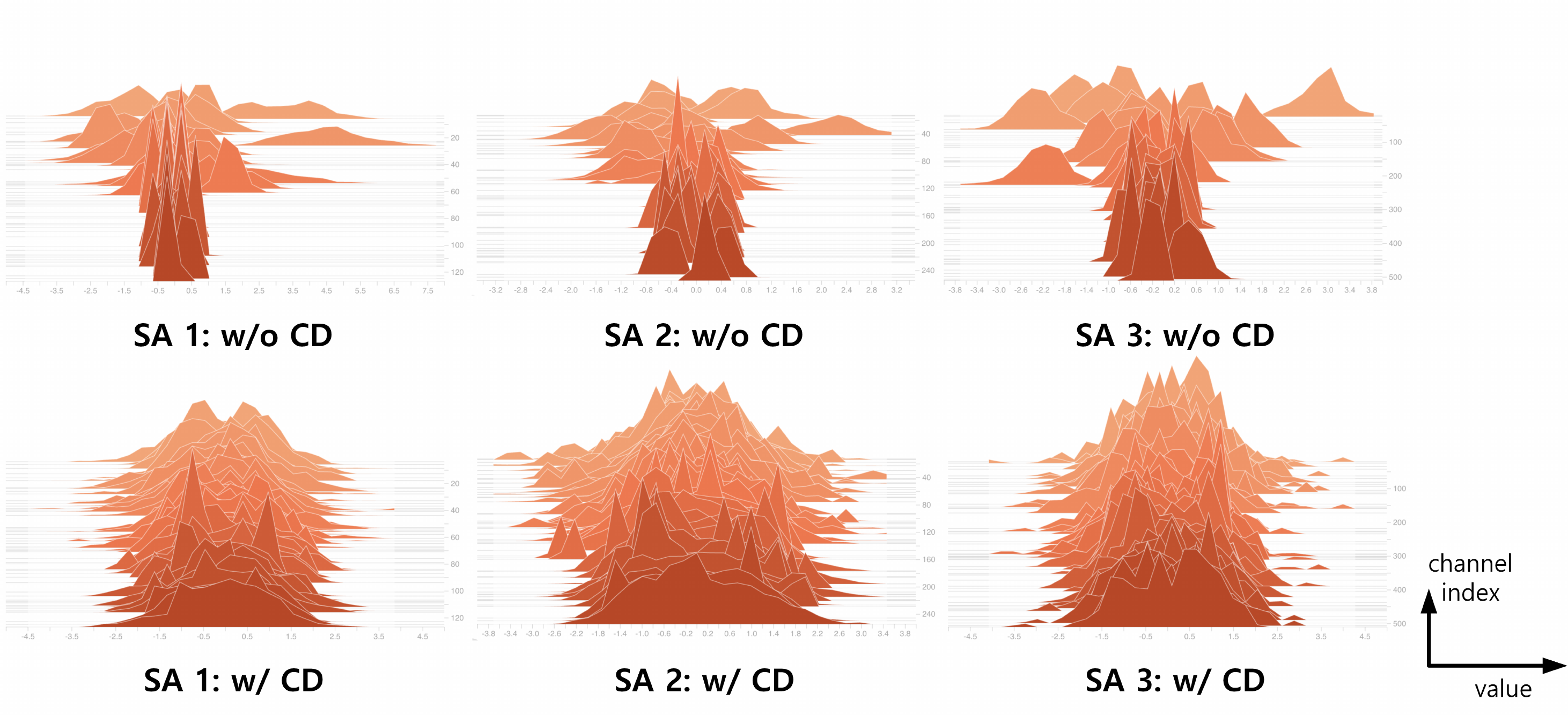}
\end{center}
   \caption{An example of the distributions of the mapped cooridnates (second-half channels, e.g., 64$\sim$128 for the left images) and the mapped features (first-half channels, e.g., 0$\sim$64 for the left images) before element-wise summation during matrix multiplication in the first layer of each stage. For an obvious comparison, we put these two modalities together in each plot, which does not mean that we perform concatenation in our CD. Note that, for the first layer of each stage, PointNet++ w/o CD performs BN \textbf{after} the summation of the mapped coordinates and features (the status like the above three images), while PointNet++ w/ CD performs BN \textbf{before} the summation (the status like the below three images). The problem of distribution imbalance will weaken the importance of one of the two kinds of input, and CD can alleviate this problem in a simple manner.}
\label{fig:cd_plot}
\end{figure*}

\section{Details of Channel De-differentiation}
\label{sec:polar}

We propose channel de-differentiation to handle the obvious distribution imbalance between the maaped cooridnates and the mapped last-stage features in each stage of set abstraction (SA) in a PointNet++ \cite{qi2017pointnet++} model. An illustration is shown in Fig.~\ref{fig:cd_plot}. This may lead to an ignorance of the input of coordinates in the last few layers of MLPs. We consider this is mainly caused by the difference of the distributions of various types of input (like coordinates and high-level features). 

Intuitively, we adopt batch normalization to alleviate the difference of these distributions. In the first MLP of each SA, the fused feature $\mathbf{f}_i^1$ of the $i$-th point can be rewrite as:
\begin{align}
\begin{split}
\mathbf{f}_i^1=\omega^1([\mathbf{x}_i, \mathbf{f}_i])=\omega_x^1(\mathbf{x}_i) + \omega_f^1(\mathbf{f}_i),
\end{split}
\end{align} 
where $\omega^1$ is a linear function, the concatenation of the weights of $\omega_x^1$ and $\omega_f^1$ equals to the weights of $\omega^1$. $\mathbf{x}_i$ and $\mathbf{f}_i$ corresponds to the coordinate and the high-level feature from the last stage of the $i$-th point, respectively.

Commonly, when we add the normalization and non-linearity to this formula, the feature can be presented as:
\begin{equation}
\mathbf{f}_i^1=ReLU(BatchNorm(\omega_x^1(\mathbf{x}_i) + \omega_f^1(\mathbf{f}_i))).
\end{equation}

Empirically, the point-based models benefit from separate application of batch normalization to $\mathbf{x}_i$ and $\mathbf{f}_i$ as follows:
\begin{align}
\begin{split}
\mathbf{f}_i^1=ReLU(BatchNorm_x(\omega_x^1(\mathbf{x}_i)) + \\
BatchNorm_f(\omega_f^1(\mathbf{f}_i))).
\end{split}
\end{align} 

\begin{figure*}
\begin{center}
    \includegraphics[width=0.9\textwidth]{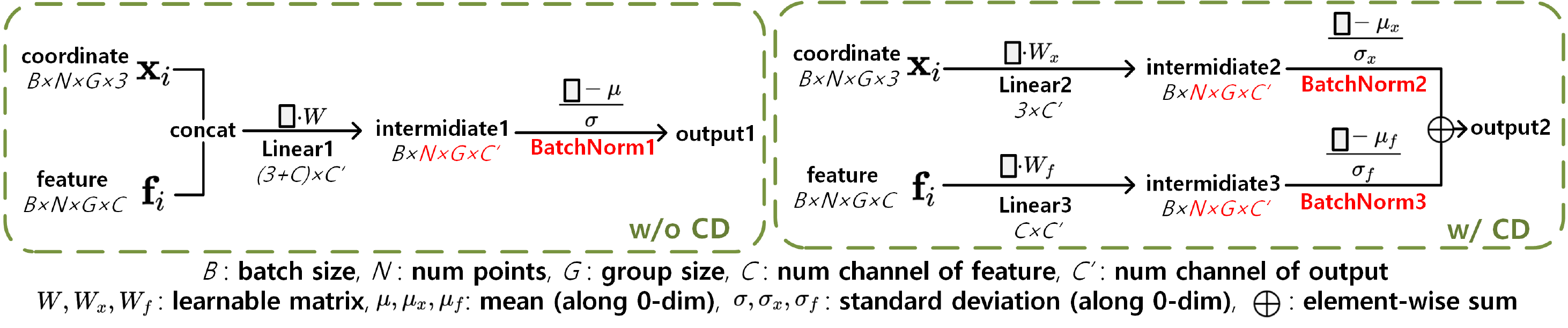}
\end{center}
   \caption{Illustration of our Channel De-differentiation.}
\label{fig:cd}
\end{figure*}

This tiny modification can significantly boost the performance of point-based models as well. For our RepSurf, $\mathbf{x}_i$ may contain polar coordinates, and $\mathbf{f}_i$ may be the features of RepSurf, RGB information. An illustration of our Channel De-differentiation is shown in Fig.~\ref{fig:cd}

\section{Computation of FLOPs}
\label{sec:flops}

To explore the efficiency of various models, we adopt the same formulas of complexity for the calculation of FLOPs. Since prior works are based on different versions of CUDA point cloud operations or non-CUDA ones, it may lead to an unfair comparison of efficiency based on FLOPs. Therefore, we treat the point cloud operations, including farthest point sampling, indexing, ball querying, knn querying, the same for the final estimation of FLOPs of different models. Following the common rules of FLOPs calculation, We count for the addition and multiplication of float points only.

For other basic operations, such as Convolution, ReLU, MLP, we adopt the default settings of THOP \footnote{\url{https://github.com/Lyken17/pytorch-OpCounter}}.

\section{Computation of Speed}
\label{sec:speed}

We test all methods with one V100 GPU and four cores Intel Xeon @ 2.50GHz CPU. The speed may vary with different sizes of input due to the parallelism of GPU. In this case, we set the batch size to 16 for all methods on the tasks of classification and segmentation. For detection, we set the batch size to 1 on the same experimental workstation in \cite{liu2021group}.

The FLOPs of one model can present the efficiency radically and theoretically. For an overall view of the efficiency, we adopt the practical method by testing the speed during the process of training and inference.

\section{Implementation details}
\label{sec:implementation}

\textbf{Classification.}~~We implement Triangular and Umbrella RepSurf on PointNet++ \cite{qi2017pointnet++} (SSG version). For both the datasets of ModelNet and ScanObjectNN, we set the initial learning rate to 0.001 with a decay rate of 0.7 for every 20 iterations. We use Adam for optimization. We apply data augmentation (including random scale, random shift, random dropout) when training on ModelNet, while we do not apply any augmentation methods for ScanObjectNN. Considering the quality of surface reconstruction, we sample 1024 points with farthest point sampling (FPS) method before input. We normalize the point clouds into the range of $[-1, 1]$ for ModelNet. We apply label smoothing with a ratio of 0.1.

\textbf{Segmentation.}~~We implement RepSurf on PointNet++ \cite{qi2017pointnet++} (SSG segmentation version). For both the datasets of S3DIS and ScanNet, we set the initial learning rate to 0.5, with a decay rate of 0.1 on the 60th and 80th iteration. We use SGD, with a weight decay of 1e$^{-4}$ for optimization. We apply data augmentation (including point cloud scaling, color contrasting, color shifting, and color jittering) when training on S3DIS and ScanNet. Considering the quality of surface reconstruction, we sample points with grid sampling method before input. We weight the loss with the ratio of classes.

\textbf{Detection.}~~We implement RepSurf on ScanNet V2 and SUN RGB-D following the practice of GroupFree \cite{liu2021group}.

\section{Detailed Experimental Results}

We reveal the details of detection on the datasets of ScanNet V2 (mAP@0.25 in Tab.~\ref{tab:scannet25} and mAP@0.5 in Tab.~\ref{tab:scannet50}) and SUN RGB-D (mAP@0.25 in Tab.~\ref{tab:sunrgbd25} and mAP@0.5 in Tab.~\ref{tab:sunrgbd50}).

\section{Visualization}
\label{sec:visual}

\subsection{Surface Reconstruction of RepSurf}
\label{sec:att}

\begin{figure}
\begin{center}
    \includegraphics[width=0.45\textwidth]{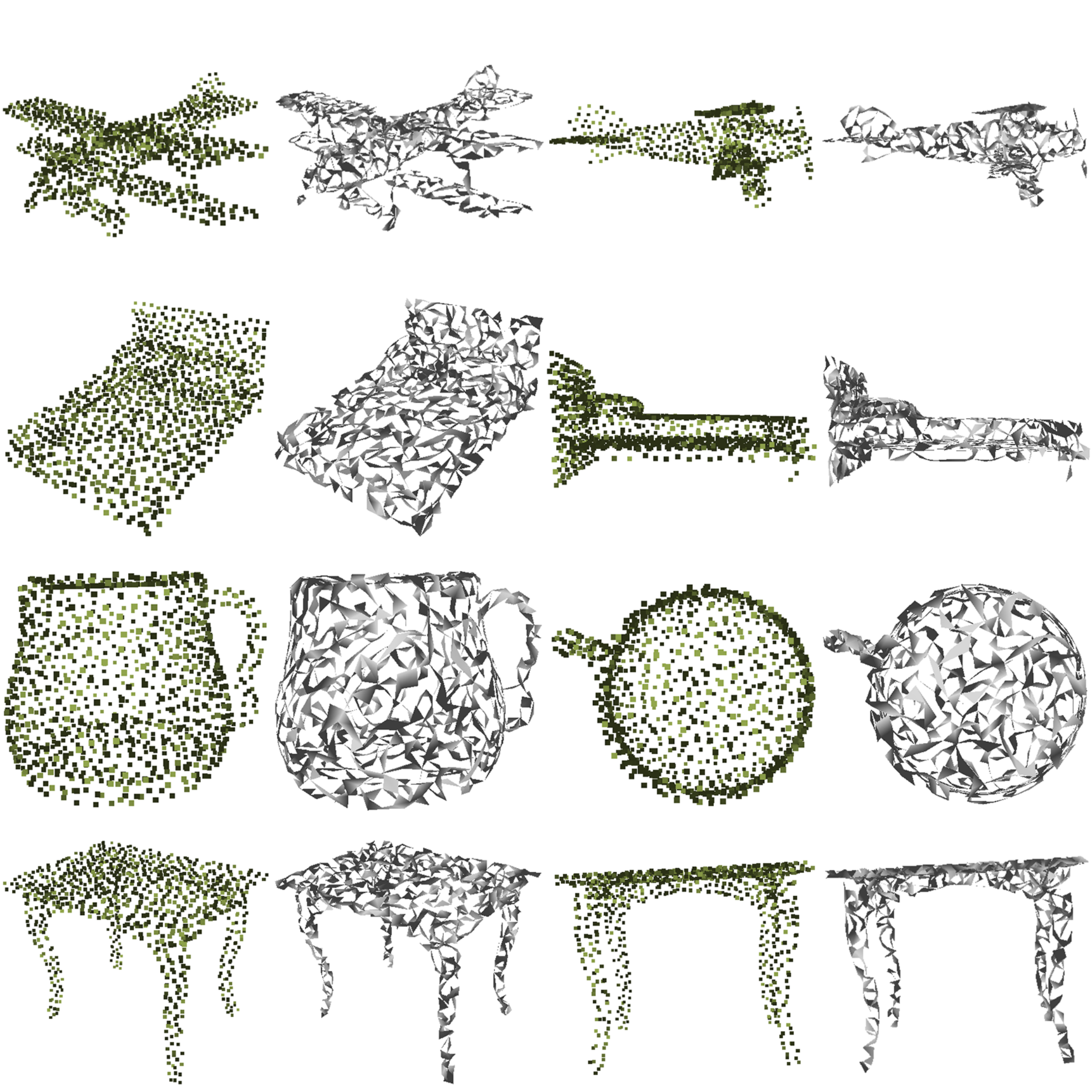}
\end{center}
   \caption{Visualization of surface reconstruction for RepSurf.}
\label{fig:surface}
\end{figure}

We visualize the results after the process of surface reconstruction in Fig.~\ref{fig:surface}. Different from prior methods, we only need to reconstruct discrete surfaces before calculating the features of Triangular and Umbrella RepSurf.

\subsection{Geometry Sensitivity on Triangular RepSurf}
\label{sec:att}

\begin{figure}
\begin{center}
    \includegraphics[width=0.45\textwidth]{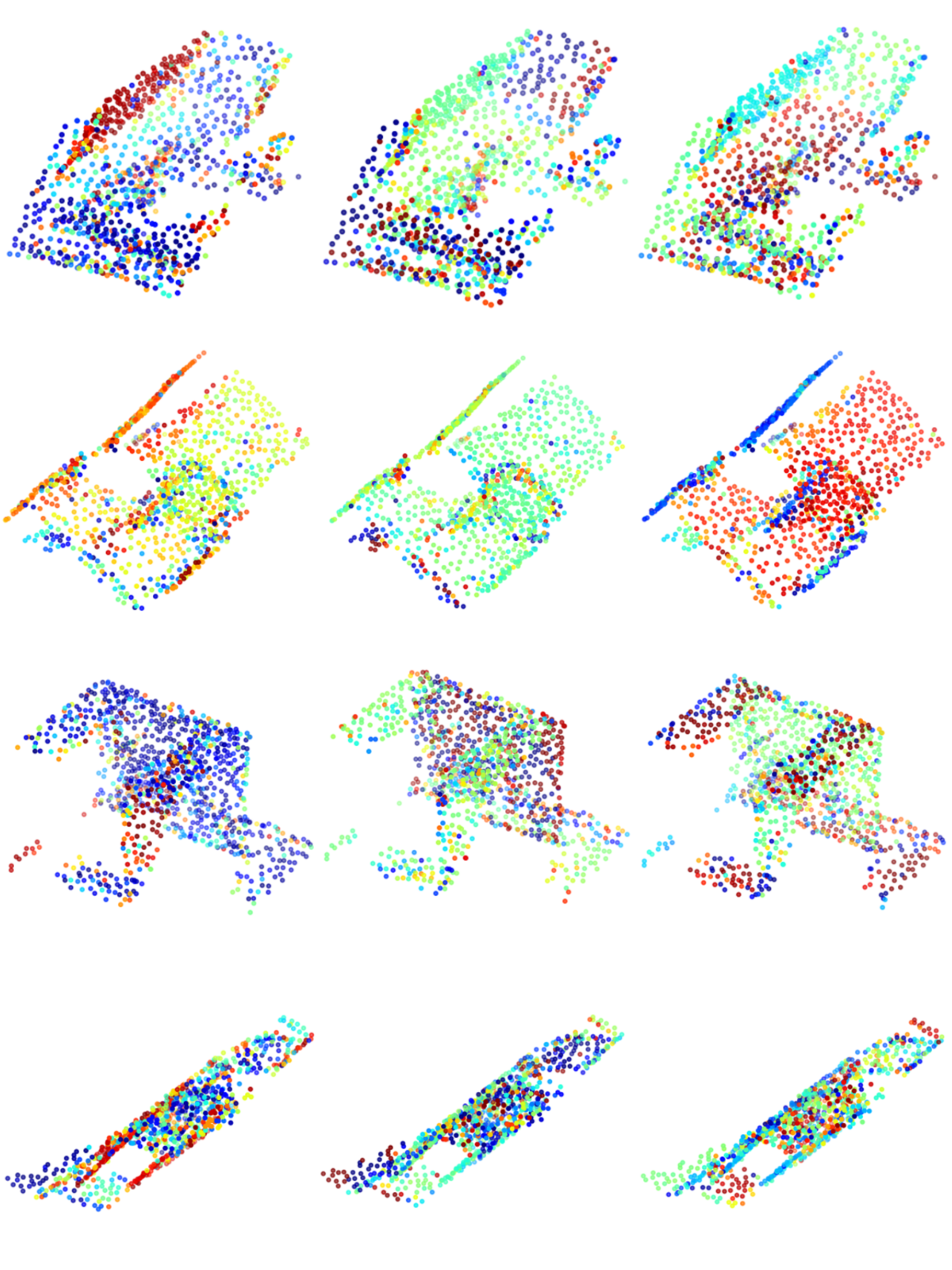}
\end{center}
   \caption{Visualization of the values of 3 channels from the normal vectors of Triangular RepSurf.}
\label{fig:tri_sense}
\end{figure}

We visualize the output of each channel of Triangular RepSurf on ScanObjectNN in Fig~\ref{fig:tri_sense}. Triangular RepSurf is eligible to perceive the local geometries numerically. Thus, the points on a flat shape have similar color, while the color of points on an edge changes obviously.

\subsection{Geometry Sensitivity on Umbrella RepSurf}
\label{sec:att}

\begin{figure*}
\begin{center}
    \includegraphics[width=\textwidth]{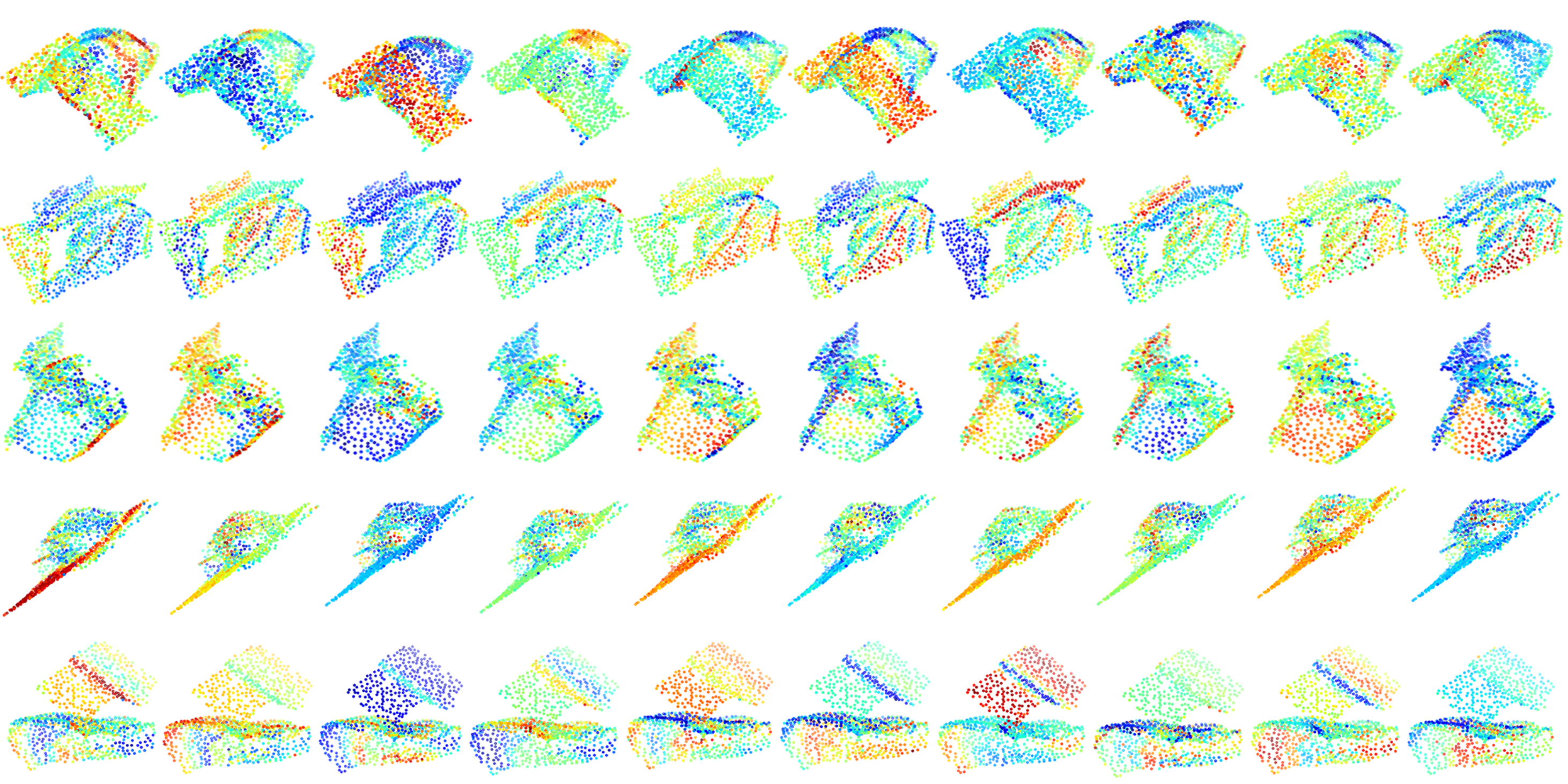}
\end{center}
   \caption{Visualization of the values of 10 channels from Umbrella RepSurf.}
\label{fig:umb_sense}
\end{figure*}

We visualize the output of each channel of Umbrella RepSurf on ScanObjectNN in Fig~\ref{fig:umb_sense}. Intuitively, Umbrella RepSurf can recognize the local geometries, including the edges and the planes of objects.

\clearpage

\begin{table*}[htb!]
\begin{center}
\footnotesize
\setlength{\tabcolsep}{2.3pt}
\begin{tabular}{l|c|cccccccccccccccccc|c}
\toprule
methods & backbone & cab & bed & chair & sofa & tabl & door & wind & bkshf & pic & cntr & desk & curt & fridg & showr & toil & sink & bath & ofurn & mAP \\ \hline
VoteNet~\cite{qi2019deep} & PointNet++ & 47.7& 88.7 & 89.5 & 89.3 & 62.1 & 54.1 & 40.8 & 54.3 & 12.0 & 63.9 & 69.4 & 52.0 & 52.5 & 73.3 & 95.9 & 52.0 & 92.5 & 42.4 & 62.9 \\
H3DNet~\cite{zhang2020h3dnet}& 4$\times$PointNet++ & 49.4 &88.6 &91.8 & 90.2 &64.9 & 61.0 &51.9& 54.9 & 18.6 &62.0 &75.9 &57.3 &57.2 &75.3 &97.9 &67.4 &92.5 &53.6 &67.2 \\ \hline
GroupFree$^{6, 256}$ & PointNet++ & 54.1 & 86.2& 92.0 & 84.8 & 67.8 & 55.8 & 46.9 & 48.5 & 15.0 & 59.4 & 80.4 & 64.2 & 57.2 & 76.3 & 97.6 &  76.8 & 92.5 & 55.0 & 67.3 \\
\rowcolor{RowColor} GroupFree$^{6, 256}$ & RepSurf-U & 55.5 & 87.7 & 93.4 & 85.9 & 69.1 & 57.3 & 48.8 & 50.0 & 16.5 & 61.0 & 81.6 & 66.2 & 59.0 & 77.5 & 99.2 & 78.2 & 94.0 & 56.8 & 68.8 \\ \hline
GroupFree$^{12, 512}$ & PointNet++$^2$ & 52.1 &  91.9 &  93.6 & 88.0 & 70.7 & 60.7 &  53.7 &  62.4 & 16.1
& 58.5 & 80.9 &  67.9 & 47.0 &  76.3 &  99.6 & 72.0 &  95.3 &56.4  &  69.1 \\
\rowcolor{RowColor} GroupFree$^{12, 512}$ & RepSurf-U$^2$ & 54.6 & 94.0 & 96.2 & 90.5 & 73.2 & 62.7 & 55.7 & 64.5 & 18.6 & 60.9 & 83.1 & 69.9 & 49.4 & 78.4 &  99.4 & 74.5 & 97.6 & 58.3 &  71.2 \\
\bottomrule
\end{tabular}
\end{center}
\caption{Performance of mAP@0.25 for each category on the ScanNet V2 dataset.}
\label{tab:scannet25}
\end{table*}

\begin{table*}[htb!]
\begin{center}
\footnotesize
\setlength{\tabcolsep}{2.3pt}
\begin{tabular}{l|c|cccccccccccccccccc|c}
\toprule
methods & backbone & cab & bed & chair & sofa & tabl & door & wind & bkshf & pic & cntr & desk & curt & fridg & showr & toil & sink & bath & ofurn & mAP \\ 
\hline
VoteNet~\cite{qi2019deep} & PointNet++ & 14.6 & 77.8 & 73.1 &  80.5 & 46.5 & 25.1 & 16.0 & 41.8 & 2.5 & 22.3 & 33.3 & 25.0 & 31.0 & 17.6 & 87.8 & 23.0 & 81.6 & 18.7 & 39.9 \\
H3DNet~\cite{zhang2020h3dnet}& 4$\times$PointNet++ & 20.5 &79.7& 80.1& 79.6& 56.2& 29.0& 21.3& 45.5 & 4.2& 33.5& 50.6& 37.3& 41.4& 37.0& 89.1& 35.1&  90.2& 35.4 & 48.1 \\
\hline
GroupFree$^{6, 256}$ & PointNet++ & 23.0 &78.4 &78.9 &68.7 &55.1 & 35.3 & 23.6 & 39.4 & 7.5 & 27.2 & 66.4 & 43.3& 43.0 & 41.2 & 89.7 & 38.0 & 83.4 & 37.3 & 48.9 \\
\rowcolor{RowColor} GroupFree$^{6, 256}$ & RepSurf-U & 24.9 & 79.6 & 80.1 & 70.4 & 56.4 & 36.7 & 25.5 & 41.4 & 8.8 & 28.7 & 68.0 & 45.2 & 45.0 & 42.7 & 91.3 & 40.1 & 85.1 & 39.2 & 50.5 \\ \hline
GroupFree$^{12, 512}$ & PointNet++$^2$ & 26.0 &  81.3 & 82.9 & 70.7 & 62.2 &  41.7 &  26.5 &  55.8 & 7.8 & 34.7 &  67.2 &  43.9&  44.3 &44.1 & 92.8 & 37.4 & 89.7 & 40.6 &  52.8 \\
\rowcolor{RowColor} GroupFree$^{12, 512}$ & RepSurf-U$^2$ & 28.5 & 83.5 & 84.8 & 72.6 & 64.0 & 43.6 & 28.3 & 57.8 & 9.6 & 37.0 & 69.7 & 45.9 & 46.4 & 46.1 & 94.9 & 39.1 & 92.1 & 42.6 &  54.8 \\

\bottomrule
\end{tabular}
\end{center}
\caption{Performance of mAP@0.5 for each category on the ScanNet V2 dataset.}
\label{tab:scannet50}
\end{table*}

\begin{table*}[htb!]
\small
\setlength{\tabcolsep}{4pt}
\begin{center}
\begin{tabular}{l|c|cccccccccc|c}
\toprule
methods & backbone & bathtub & bed & bkshf & chair & desk & drser & nigtstd & sofa & table & toilet & mAP \\
\hline
VoteNet~\cite{qi2019deep} & PointNet++& 75.5 & 85.6 & 31.9 & 77.4 & 24.8 & 27.9 & 58.6 & 67.4 & 51.1 & 90.5 & 59.1\\
H3DNet~\cite{zhang2020h3dnet}& 4$\times$PointNet++ & 73.8 &85.6 &31.0& 76.7 &29.6& 33.4& 65.5& 66.5& 50.8& 88.2& 60.1 \\
GroupFree$^{6, 256}$ & PointNet++ &  80.0 &  87.8 & 32.5  &  79.4 &32.6 &  36.0 &  66.7  &  70.0  &  53.8  & 91.1  &  63.0\\ \hline
\rowcolor{RowColor} GroupFree$^{6, 256}$ & RepSurf-U &  81.1 & 89.3 & 34.4 & 80.4 & 33.5 & 37.3 & 68.1 & 71.4 & 54.8 & 92.3  &  64.3\\ 
\rowcolor{RowColor} GroupFree$^{12, 256}$ & RepSurf-U$^2$ & 81.9 & 89.9 & 35.3 & 81.2 & 33.5 & 38.1 & 68.8 & 71.5 & 55.6 & 93.2  &  64.9\\ 
\bottomrule
\end{tabular}
\end{center}
\caption{Performance of mAP@0.25 for each category on the SUN RGB-D validation set.}
\label{tab:sunrgbd25}
\end{table*}

\begin{table*}[htb!]
\small
\setlength{\tabcolsep}{4pt}
\begin{center}
\begin{tabular}{l|c|cccccccccc|c}
\toprule
methods & backbone & bathtub & bed & bkshf & chair & desk & drser & nigtstd & sofa & table & toilet & mAP \\
\hline
VoteNet~\cite{qi2019deep} & PointNet++ & 45.4 & 53.4 & 6.8 & 56.5 & 5.9 & 12.0 & 38.6 & 49.1 & 21.3 & 68.5 & 35.8 \\
H3DNet~\cite{zhang2020h3dnet}& 4$\times$PointNet++ & 47.6 &52.9 &8.6& 60.1 & 8.4& 20.6& 45.6& 50.4& 27.1& 69.1& 39.0 \\
GroupFree$^{6, 256}$ & PointNet++ &  64.0 &  67.1 &  12.4 &  62.6 & 14.5 &  21.9&  49.8&  58.2 &  29.2 &  72.2 &  45.2\\ \hline
\rowcolor{RowColor} GroupFree$^{6, 256}$ & RepSurf-U &  65.2 & 67.5 & 13.2 & 63.4 & 15.0 & 22.4 & 50.9 & 58.8 & 30.0 & 72.7 &  45.9\\
\rowcolor{RowColor} GroupFree$^{12, 512}$ & RepSurf-U$^2$ & 66.5 & 70.0 & 14.9 & 64.7 & 17.0 & 24.7 & 52.0 & 60.7 & 31.7 & 74.4 &  47.7\\
\bottomrule
\end{tabular}
\end{center}
\caption{Performance of mAP@0.5 for each category on the SUN RGB-D validation set.} 
\label{tab:sunrgbd50}
\end{table*}

\end{document}